\documentclass{article}

\PassOptionsToPackage{numbers, sort&compress}{natbib}
\usepackage[preprint]{neurips_2026}

\usepackage[utf8]{inputenc} 
\usepackage[T1]{fontenc}    
\usepackage{hyperref}       
\usepackage{url}            
\usepackage{booktabs}       
\usepackage{amsfonts}       
\usepackage{nicefrac}       
\usepackage{microtype}      
\usepackage{xcolor}         
\usepackage{mathtools}
\usepackage{amsthm}
\usepackage{adjustbox}
\usepackage{comment}
\usepackage{enumitem}
\usepackage{multirow}
\usepackage[capitalize,noabbrev]{cleveref}
\usepackage{caption}
\usepackage[hang,flushmargin]{footmisc}

\hypersetup{
    colorlinks=true,
    linkcolor=black,
    citecolor=black,
    urlcolor=black
}
 
\title{Birds of a Feather Flock Together: Background-Invariant Representations via Linear Structure in VLMs}

\author{
  Youssef Zaazou \\
  Independent Researcher \\
  \texttt{yazaazou@mun.ca}
\And
  Mark Thomas \\
  Independent Researcher \\
  \texttt{mark.thomas@jasco.com}
}

\newcommand{\std}[1]{\scalebox{0.7}{{ $\pm #1$}}}

\begin{document}

\maketitle

\begin{abstract}

Vision-language models (VLMs), such as CLIP and SigLIP 2, are widely used for image classification, yet their vision encoders remain vulnerable to systematic biases that undermine robustness. In particular, correlations between foreground objects and their backgrounds constitute a salient and practically important class of spurious dependencies. In this work, we revisit the well-known property of high linear additivity in VLM embedding spaces and show that it enables a decomposition of scene representations into foreground and background components. Leveraging this insight, we introduce a pre-training approach that exploits this property to construct background-invariant representations using synthetic data. Our method achieves, to our knowledge, the first worst-group accuracy exceeding $90\%$ on Waterbirds under perfect ($100\%$) spurious correlation (i.e., no minority-group examples in the training data). Furthermore, it demonstrates strong sim-to-real transfer and requires no access to real-world debiased data, making it practical for real-world deployment.
\end{abstract}

\section{Introduction}\label{sec:intro}

Vision-language models (VLMs), such as CLIP \cite{CLIP_original} and SigLIP 2 \citep{siglip_2}, have become widely adopted for image classification, both in zero-shot settings \citep{CLIP_original, siglip_2} and as frozen feature extractors paired with lightweight heads or adapters  \cite{clip_adapt}.  Their flexibility and strong transfer performance have led to their use in a wide range of downstream tasks such as recognition, detection, retrieval, and multimodal reasoning \cite{transformer_survey, clip_survey}. 

However, VLM vision encoders exhibit notable biases and sensitivity to spurious signals \citep{clip_bias_1, sober_clip, varma2024ravl, BEE,janouskova2025robust}. A particular source of errors, which has significant practical downsides, is background spurious correlations. Backgrounds can often encode strong but coincidental links with foreground objects. Of high practical concern are spurious background correlations in medical imaging, where certain diseases may be misclassified due to erroneous links to specific scanners or acquisition protocols \citep{medical_one, medical_two}. Furthermore, there is evidence that contrastive training encourages the image encoder to associate foreground and background signals which hurts generalizability and cross-domain stability \citep{sober_clip, foreground_or_background}. 

In this work, we examine VLM background bias by revisiting the embedding space property of high linear additivity \citep{linear_CLIP, linear_VLM1,linear_VLM2}, and show that VLM representations exhibit strong linear additivity between foreground and background concepts. We exploit this insight to introduce Background-invariant Anchor Pre-training (BAP), a novel approach leveraging synthetic data to robustify vision encoders before they are exposed to target downstream distributions. Our primary contributions are summarized as follows:

\begin{itemize}[leftmargin=20pt,topsep=0pt,itemsep=1pt]
    
    \item We introduce a two-stage pre-training method that robustifies VLM vision encoders by exploiting high linear additivity between foreground and background representations to produce background-invariant representations.
    
    \item  We achieve state-of-the-art worst-group accuracy (WGA) exceeding $90\%$ on Waterbirds under perfect spurious correlation, where no minority-group examples are present in training or validation sets. We further show that BAP transfers to real-world data and does not overfit to segmentation artifacts despite relying on synthetic data.
    
    \item  We demonstrate that BAP-induced robustness generalizes to the super-class level. Unlike previous methods that require re-optimization for each downstream binary task (e.g., Car vs. Truck, Bike vs. Motorbike), BAP is applied only once per super-class (e.g., Vehicles). Once pre-trained, the same feature extractor can be reused for multiple downstream tasks with minimal deployment constraints.
    
    \item We demonstrate high practicality: BAP achieves near-peak robustness using as few as $50$ segmented foreground items, and remains resilient to coarse or imperfect segmentation masks (e.g., bounding boxes).
\end{itemize}

To cover the contrastive VLM landscape, we evaluate both CLIP \cite{CLIP_original} and SigLIP 2 \citep{siglip_2}, representing the foundational Softmax-based baseline and the state-of-the-art Sigmoid-based architecture, respectively. The code to reproduce our results is available at our
\href{https://github.com/BAP-user/BAP_reproducibility_repo}{GitHub repository}.

\section{Motivation}\label{sec:image encoder additivity}

Prior work establishes that VLM embedding spaces (e.g., CLIP, SigLIP2) exhibit a high degree of linear additivity and structural compositionality \citep{linear_CLIP, linear_VLM1, linear_VLM2}. We begin by reformulating this property through the lens of foreground-background compositionality specifically. We test whether scene representations behave as a linear superposition of separate object and background components.  For each paired object image $I_a$, background image $I_b$, and composite scene $I_{a,b}$, a vision encoder $g(\cdot)$ produces unit-norm embeddings:
\[
v_a = g(I_a), \quad v_b = g(I_b), \quad v_{a,b} = g(I_{a,b}).
\]

 We quantify a vision encoder's tendency to represent backgrounds and foregrounds independently by measuring the cosine similarity between the composite embedding and the vector sum of its parts:
\begin{equation}
\text{Sim}(I_a, I_b ; I_{a,b})
=
\frac{v_{a,b} \cdot (v_a + v_b)}{\|v_{a,b}\| \, \|v_a + v_b\|}.
\end{equation}\label{eq:both_sum}

We define the linear additivity score as $S=\text{Sim}(I_a,I_b;I_{a,b})$, where values near $1.0$ indicate that the scene is encoded as a near-linear superposition of its constituent components. We evaluate Equation \ref{eq:both_sum} on 300,000 scene composites $(I_{a,b})$ and their constituents $(I_a,I_b)$, constructed from random MS-COCO objects paired with random Places365 backgrounds \citep{coco,places}. Specifically, we compare CLIP and SigLIP2 vision encoders against supervised vision encoders (ImageNet-1K/21K) and self-supervised (SSL) representatives like MAE and DINOv2 \citep{IN1K, IN21k, DinoV2,MAE}. Finally, we evaluate both ViT and ConvNeXt backbones \citep{vit, convnext}, with results reported in Table \ref{tab:image_additivity}. To ensure parity with VLM designs, all embeddings are unit-normalized for a fair comparison.

\begin{table}[h]

\caption{$S$ scores (mean $\pm$ std over 300,000 composites) across VLM, supervised, and SSL backbones; as $S$ approaches 1, the degree of linear additivity between foreground and background representations increases. Elevated values of $S$ indicate a high level of linear additivity in VLM embeddings.
}
\label{tab:image_additivity}
\centering

\setlength{\tabcolsep}{5pt} 
\begin{tabular}{@{}l cc cc cc@{}}
\toprule
& \multicolumn{2}{c}{\textbf{VLM}} & \multicolumn{2}{c}{\textbf{Supervised}} & \multicolumn{2}{c}{\textbf{SSL}} \\
\cmidrule(lr){2-3} \cmidrule(lr){4-5} \cmidrule(lr){6-7}
\textbf{Architecture} & \textbf{SigLIP 2}$^{\dagger}$ & \textbf{CLIP} & \textbf{IN-1K}$^{\dagger}$ & \textbf{IN-21K} & \textbf{DINOv2} & \textbf{MAE}$^{\dagger}$ \\
\midrule
ViT-B/16   & 0.91 \std{0.02} & 0.82 \std{0.04} & 0.72 \std{0.04} & 0.75 \std{0.03} & 0.66 \std{0.05} & 0.73 \std{0.06} \\
ViT-L/14$^{\dagger}$  & 0.90 \std{0.03} & 0.81 \std{0.03} & 0.73 \std{0.04} & 0.74 \std{0.05} & 0.64 \std{0.07} & 0.72  \std{0.05}  \\
ConvNeXT$^{\ddagger}$& --             & 0.79 \std{0.04} & 0.66 \std{0.07} & 0.63 \std{0.11} & --  & 0.71 \std{0.05} \\
\bottomrule
\end{tabular}
\end{table}

\renewcommand{\thefootnote}{\fnsymbol{footnote}}

\footnotetext[2]{Note the SigLIP 2, IN-1K, and MAE models utilize a ViT-L/16 backbone  rather than a ViT-L/14.} 
\footnotetext[3]{ Unfortunately, no pre-trained checkpoints exist for SigLIP 2 and DINOv2 using a ConvNeXT backbone.}   

The observations in Table \ref{tab:image_additivity} confirm that the high linear additivity observed in VLM embedding spaces \citep{linear_CLIP,linear_VLM1,linear_VLM2} extends to foreground and background representations. Perhaps most surprising are the $S$ values for SigLIP2, which  are substantially higher than corresponding supervised and self-supervised models. Overall, the elevated levels of linear additivity hold for both SigLIP2 and CLIP across a variety of backbones; consequently, background components remain strongly encoded in the embedding space which may account for the background-based spurious errors prevalent in these models. 

Our findings are further replicated in VLM text encoders, which behave similarly to `bag-of-words' \citep{bag} models such as Word2Vec \citep{word2vec} rather than contextual transformer models such as BERT \citep{BERT}. This indicates a consistent behavior across both modalities; results for the text-encoder analyses are provided in Appendix \ref{app: text encoder addititivty}.

\section{Methodology}\label{sec: methodology}

We reformulate the mitigation of spurious correlations as a robustness pre-training stage, departing from downstream or post-hoc interventions \citep{dfr, afr, dial, roboshot}. Encoding robustness directly into the representation prior to task-specific fine-tuning secures the primary advantages outlined in Section \ref{sec:intro}: super-class task-agnostic robustness and resilience to perfect ($100\%$) spurious correlations.

\subsection{The BAP algorithm}\label{sec: BAP algorithm}
Our method is designed to take advantage of the linear additivity property shown in Table \ref{tab:image_additivity}. BAP is composed of two sequential phases as can be seen in Figure \ref{fig:bap_flowchart}. First we extract a set of object-specific, foreground-only anchor vectors from the vision encoder's embedding space using a frozen, contrastively pre-trained teacher. Second, we optimize the student model to map the object, appearing in numerous novel, randomized contexts, to this invariant anchor vector. By doing so, we distill background-invariance directly into the student's representation space.

\subsubsection{Phase 1: anchor vector extraction}\label{sec: BAP phase 1}

We begin by constructing a dictionary of foreground-only anchor vectors (training target vectors) used during the robustness pre-training stage (each foreground object gets its own anchor vector). We utilize a frozen vision encoder, $ f_{\theta}^*$, as a teacher model to generate these anchors. For a specific foreground object instance ($x_{fg}$) we generate a set of composites by superimposing the object onto diverse, randomly sampled background images, $b_k$. We compute the normalized feature embedding ($\bar{\mathbf{z}}$) for each composite and derive the anchor ($\mathbf{a}$) by averaging these vectors and normalizing:

\begin{equation}\label{eq:anchor avg}
    \bar{\mathbf{z}} = \frac{1}{K} \sum_{k=1}^{K} f_{\theta}^*(\mathcal{C}(x_{fg}, b_k)),  \qquad  \mathbf{a} = \frac{\bar{\mathbf{z}}}{\| \bar{\mathbf{z}} \|_2},
\end{equation}

where $\mathcal{C}$ denotes the compositing function.

\subsubsection{Phase 2: Robust alignment pre-Training}\label{sec: BAP phase 2}

For BAP's second phase, we unfreeze the vision encoder ($f_{\theta}$) and optimize it to align its embeddings with the pre-computed anchors. During this phase, each foreground instance is composited onto $M$ random backgrounds to generate $M$ composites, denoted as $\{\hat{x}_m\}_{m=1}^M$. The model is then optimized to minimize the cosine distance between the embedding of each randomized input $\hat{x}_m$ and the corresponding anchor vector ($\mathbf{a}$):

\begin{equation}
\mathcal{L}_{align} = 1 - \frac{f_{\theta}(\hat{x}_m)^\top \mathbf{a}}{\|f_{\theta}(\hat{x}_m)\|_2 \|\mathbf{a}\|_2}, \quad m \in \{1, \dots, M\}
\label{eq:align loss}
\end{equation}

\subsection{Algorithmic justification}\label{sec: algo}

\paragraph{Phase 1: anchor vector derivation and properties}

Our motivation for Equation \ref{eq:anchor avg} stems from the observations in Table \ref{tab:image_additivity} regarding the high level of linear additivity in VLM embedding spaces. Based on this property, an embedding vector $\mathbf{z}$ may be approximated as the linear superposition of foreground ($\mathbf{v}_{fg}$) and background ($\mathbf{v}_{bg}$) component vectors:

\begin{equation}\label{eq:linear_anchor}
       \mathbf{z} \approx \mathbf{v}_{fg} + \mathbf{v}_{bg}.
\end{equation}

While Equation \ref{eq:linear_anchor} is an approximation ($S < 1.0$), we hypothesize that contextual residuals vanish through randomized averaging. After applying Equation \ref{eq:anchor avg}, our anchor $\mathbf{a}$ averages $K$ diverse composites from a broad distribution (e.g., Places365) allowing us to write:

\begin{equation}\label{eq:a_decomp}
    \mathbf{a} \approx \mathbf{v}_{fg} + \frac{1}{K} \sum_{k=1}^{K} \mathbf{v}_{bg}^{(k)}.
    \end{equation}

By the Law of Large Numbers, as $K$ increases, the sample mean of the backgrounds converges to its expected value $\boldsymbol{\mu}_{bg} = \mathbb{E}[\mathbf{v}_{bg}]$. Because backgrounds are sampled randomly and independently of the foreground class, $\boldsymbol{\mu}_{bg}$ is strictly \textbf{class-agnostic} and provides no discriminative information for downstream classification. Furthermore, VLM embeddings exist on a high-dimensional unit hypersphere, but natural images only occupy a narrow cone within it \citep{linear_VLM2}. Consequently, the centroid of all backgrounds ($\boldsymbol{\mu}_{bg}$) is a strictly non-zero vector pointing toward this general image manifold. For a finite $K$, the anchor retains residual noise, defined as:

\begin{equation}\label{eq:epsilon}
    \boldsymbol{\epsilon} = \left( \frac{1}{K} \sum_{k=1}^{K} \mathbf{v}_{bg}^{(k)} \right) - \boldsymbol{\mu}_{bg}
    \end{equation}

\begin{figure}[h]
    \centering
    \includegraphics[trim={0.25cm} {0.3cm} {0.25cm} {0.55cm}, clip,width=\linewidth]{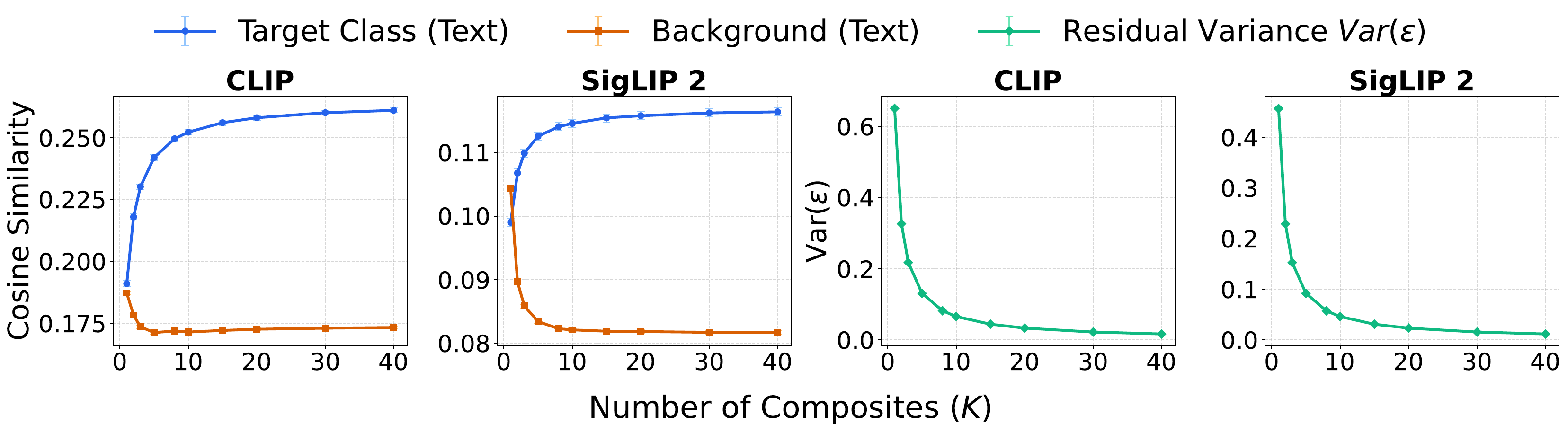}
    \caption{\textbf{Ablation on $K$.} \textbf{Left:} Anchor vector cosine similarity to foreground vs.\ background text prompts indicating that the application of Equation \ref{eq:anchor avg} preserves foreground signals while suppressing background signals. \textbf{Right:} Residual background variance $Var(\boldsymbol{\epsilon})$. As $K$ increases, $Var(\boldsymbol{\epsilon})$ exhibits a strict $1/K$ decay, confirming the relationship in Equation \ref{eq:epsilon}.}
    \label{fig:k_ablation}
\end{figure}

\begin{figure*}[t]
    \centering
    \begin{minipage}{\linewidth}
        \centering
        \includegraphics[trim={0.2cm} {0.1cm} {0.5cm} {0.4cm}, clip,width=\linewidth]{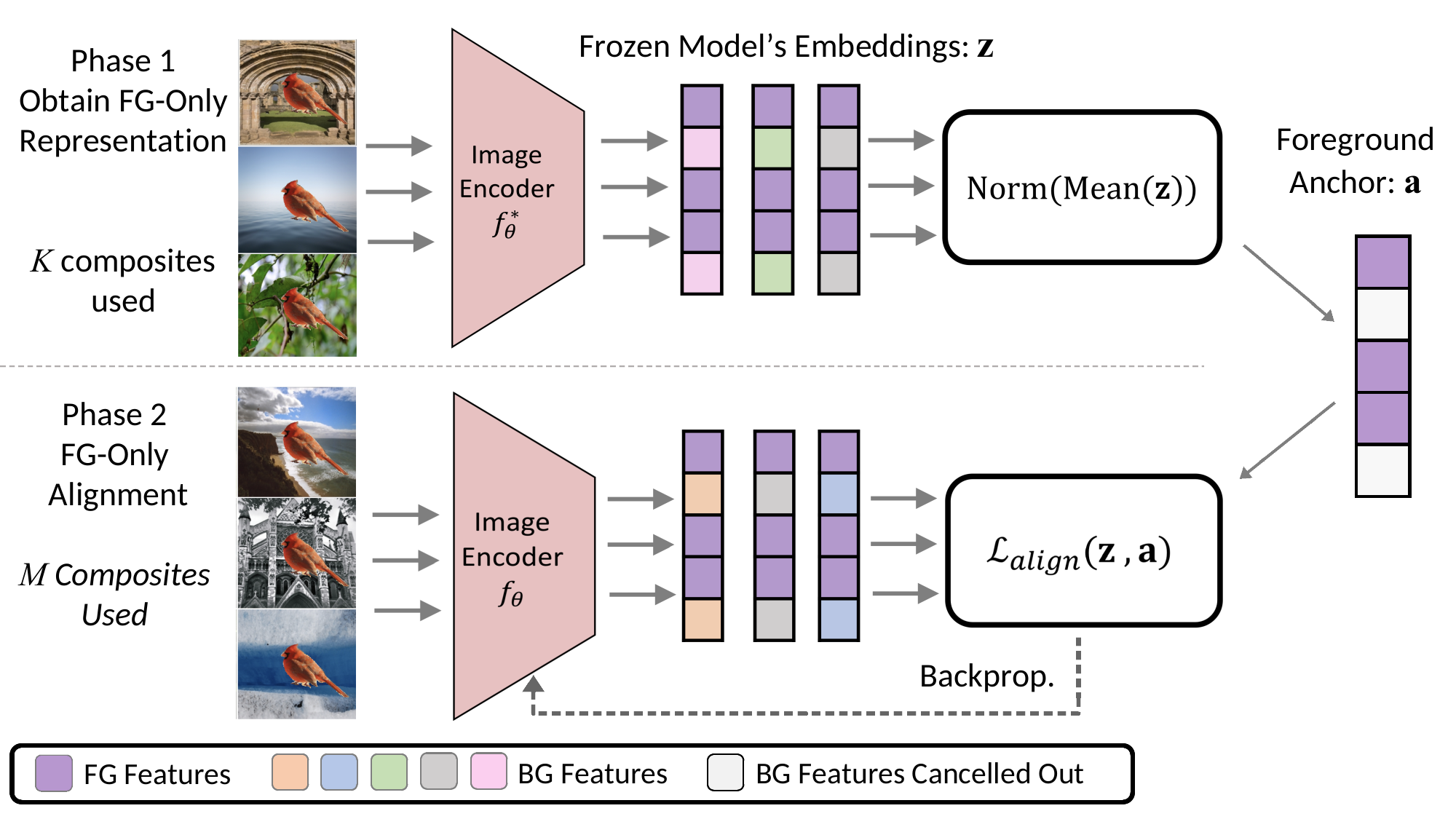}

\caption{\textbf{Overview of the BAP Pipeline.} The method consists of two sequential phases. \textbf{Phase 1 (Top): Anchor Extraction.} A foreground-specific anchor $\mathbf{a}$ is generated by averaging the embeddings $\mathbf{z}$ of a fixed object composited onto $K$ randomized backgrounds using a frozen teacher model $f_{\theta}^*$. \textbf{Phase 2 (Bottom): Robust Alignment.} The student encoder $f_{\theta}$ is optimized via $\mathcal{L}_{align}$ to map the same object across $M$ novel contexts to the invariant anchor $\mathbf{a}$. \textit{Feature Legend:} Purple slots represent core foreground features, white slots indicate suppressed background signals, and other colors denote randomized background noise.}
\label{fig:bap_flowchart}
            
\end{minipage}
\end{figure*}

Under the assumption that the background samples are independent and identically distributed (i.i.d.), the variance of this residual noise scales inversely with $K$ as $Var(\boldsymbol{\epsilon}) \propto 1/K$. We empirically confirm this theoretical $1/K$ decay in Figure \ref{fig:k_ablation} by measuring the variance of $\boldsymbol{\epsilon}$ against a true population mean ($\boldsymbol{\mu}_{bg}$) estimated from 100,000 background samples. To validate that the mathematical postulations in Equations \ref{eq:a_decomp} and \ref{eq:epsilon} hold for finite $K$ and that our anchor vector construction (Equation \ref{eq:anchor avg}) retains foreground signals while attenuating any discriminative background signals, we present the following ablation on the $K$ hyperparameter. 

Anchor vectors are generated by compositing segmented target objects (MS-COCO objects) onto $K$ diverse, randomized backgrounds from Places365. After processing the embeddings via Equation (2), we quantify signal isolation by tracking the anchor's cosine similarity against the text embeddings of both the true foreground class and the background category as $K$ increases (we record the maximum similarity across background categories; full methodology may be seen in Appendix \ref{app:k_ablation_appendix}).

Figure \ref{fig:k_ablation} confirms our mathematical motivation. The increase in cosine similarity with foreground text embeddings as $K$ increases indicates that our foreground component $\mathbf{v}_{fg}$ in Equation \ref{eq:linear_anchor} is retained. Meanwhile, the corresponding decrease in similarity with the background text embedding supports the claim that as $K$ increases, the background component and context-dependent interaction terms tend towards the non-zero offset vector ($\boldsymbol{\mu}_{bg}$) pointing towards the general manifold of background embeddings. Again, $\boldsymbol{\mu}_{bg}$ carries no  discriminative signal for downstream classifiers.

\paragraph{Phase 2: alignment via many-to-one mapping}
Now that we have established that our anchor vector generation retains foreground information while suppressing discriminative background information, we proceed to justify phase 2 of BAP. To empirically validate that the $\mathcal{L}_{align}$ objective actively suppresses background entanglement, we investigate its function as a representational bottleneck. By analogy to a $\beta$-VAE \citep{bvae}, enforcing a strict many-to-one mapping (where $M$ distinct background variations of an object are forced to map to a singular point $\mathbf{a}$) constrains the encoder to discard high-variance, unshared features (the backgrounds) and encode only the shared, salient factors (the foreground object).  

To test this, we ablate the Phase 1 anchor generation entirely. We replace the instance-specific semantic anchors with two random, orthogonal unit vectors, assigning all composites of a given class to one vector and the rest to the other (e.g., waterbirds vs. landbirds). When trained via $\mathcal{L}_{align}$ on these two static targets, the model still achieves a $90.2\%$ Worst-Group Accuracy on an in-distribution synthetic task (detailed in Section \ref{sec: results}), confirming that the many-to-one alignment objective alone is sufficient to bottleneck and suppress background signals. However, performance on real-world out-of-distribution tasks collapses catastrophically (e.g., $11.7\%$ WGA). This validates our two-phase design: Phase 2 ($\mathcal{L}_{align}$) provides the geometric bottleneck required to scrub spurious backgrounds, while Phase 1 (Teacher-distilled anchors) is strictly necessary to preserve the rich semantic structure required to prevent catastrophic forgetting and enable out-of-distribution generalization. A more in-depth version of this ablation may be seen in Appendix \ref{app: orthogonal targets}.

\section{Results}\label{sec: results}

\paragraph{Datasets and model selection}

We evaluate BAP across a diverse suite of benchmarks designed to test specific facets of robustness. We use Waterbirds for our initial results alongside CUB-200-2011 and Places365 to construct our synthetic training data \citep{waterbirds, cub,places}. Furthermore, we utilize instances from MS-COCO \cite{coco} to perform gender classification where each gender is placed in stereotypical environments (women in kitchen and men in construction fields). We demonstrate that BAP reduces reliance on spurious background cues that may encode demographic biases. Since our gender classification dataset is a custom-built dataset, we provide all details related to the training and test data construction in Appendix \ref{sec:appendix_coco_gender}. For both setups, we show results using both $95\%$ and $100\%$ spurious correlation rates in the training setup but maintain identical test sets with balanced 4-way splits across both domains to ensure a fair comparison.

While the above two evaluations use synthetic data, we use two \textbf{real-world datasets to demonstrate generalizability beyond synthetic data}. We test BAP on CounterAnimal \cite{sober_clip} with CUB-200-2011 pre-training, and on NICO++ \citep{nico++} with MS-COCO pre-training, to assess OOD generalization across varied domains and real-world contexts. This evaluation comprises a series of downstream binary vehicle classification tasks; NICO++ presents a significant challenge due to the prevalence of truncated, occluded, or incomplete object instances. The pre-training data is comprised of vehicle instances from the MS-COCO dataset; however, we exclude 'truck' and 'motorbike' instances from the pre-training set to evaluate whether BAP modifies the inductive bias at the super-class level (vehicles). Both these contexts are tested using a $100\%$ spurious correlation rate in the training data to stress test BAP (e.g., for the NICO++ evaluations, all cars appear on grass while all trucks appear on water).

We evaluate on CLIP and SigLIP 2 to bracket dual-encoder VLMs across the two dominant contrastive paradigms: Softmax-based (CLIP) and Sigmoid-based (SigLIP 2). We use a ViT-B/16 backbone for both CLIP and SigLIP 2 across all experiments, except for NICO++ where we employ a CLIP ConvNext-W backbone to demonstrate cross-architectural utility. CLIP models are initialized from LAION \citep{laion} pre-trained checkpoints. Additional results using CLIP ConvNeXT-W on Waterbirds and SigLIP 2 on NICO++ are provided in Appendix \ref{app:extra_results}. Furthermore, ablations on backbone size and non-LAION pre-trained CLIP backbones are also included in the Appendix \ref{app:backbone_ablation}.

\paragraph{Baselines and evaluation methodology} 

As a representation-level intervention, BAP is designed to yield a robust, general-purpose feature extractor within a given super-class. Consequently, our primary evaluation utilizes linear probing (LP), where the frozen BAP backbone is paired with task-specific classifiers trained via standard ERM on downstream datasets (e.g., Waterbirds). This demonstrates BAP’s core utility: a single pre-trained backbone can seamlessly support multiple downstream tasks. We additionally report zero-shot (ZS) performance. Evaluating the model without a linear head allows us to verify that BAP successfully preserves native multimodal alignment and that robustness gains stem directly from our unsupervised background suppression rather than linear probe artifacts. Because zero-shot evaluations do not utilize the downstream training sets, their reported performance is naturally identical across varying spurious correlation splits (e.g., $95\%$ vs. $100\%$)

A critical confounder of BAP's pre-training setup is data exposure; simply fine-tuning the model on a large variety of objects on randomized backgrounds might improve robustness regardless of the alignment objective. To causally isolate the efficacy of BAP's pre-training logic, we introduce a \textbf{Data-Matched Control} which undergoes Phase 1 training using the exact same randomly generated composite images and the exact same compute budget as BAP, but is optimized via standard Cross-Entropy allowing us isolate BAP's algorithmic contribution from mere increased data exposure. Crucially, this control possesses an inherent advantage over BAP: it is explicitly trained using ground-truth labels to classify the specific downstream target classes. Additionally, we evaluate BAP against a comprehensive suite of general and VLM-specific robustness baselines, spanning last-layer retraining \citep{dfr, afr}, weight-space ensembling \cite{wiseft}, and state-of-the-art representation-level interventions \citep{prusc, roboshot, dial}. Complete descriptions of these baselines is provided in Appendix \ref{app: baselines}.

\subsection{Initial findings: Waterbirds and COCO gender classification}\label{sec: waterbirds results}

\begin{table}[ht]
  \caption{Average and worst-group accuracies (mean $\pm$ std over 5 runs) across various baselines for the Waterbirds benchmark using 95\% and 100\%  spurious correlation rates. We omit the last-layer retraining baselines (DFR, AFR) for the $100\%$ benchmarks since it is established \citep{waterbirds100} that they fail in the absence of minority group data. (Note: Zero-shot evaluations report identical results across both domains as they do not utilize the correlated training sets). }
  \label{tab:waterbirds}
  \centering
  \setlength{\tabcolsep}{2.5pt} 
  \begin{small} 
  \begin{tabular}{l cccc cccc}
  \toprule
  & \multicolumn{4}{c}{\textbf{CLIP}} & \multicolumn{4}{c}{\textbf{SigLIP2}} \\
  \cmidrule(lr){2-5} \cmidrule(lr){6-9}
  & \multicolumn{2}{c}{\textbf{Waterbirds-95\%}} & \multicolumn{2}{c}{\textbf{Waterbirds-100\%}} & \multicolumn{2}{c}{\textbf{Waterbirds-95\%}} & \multicolumn{2}{c}{\textbf{Waterbirds-100\%}} \\
  \cmidrule(lr){2-3} \cmidrule(lr){4-5} \cmidrule(lr){6-7} \cmidrule(lr){8-9}
  \textbf{Method} & \textbf{AVG} ($\uparrow$) & \textbf{WGA} ($\uparrow$) & \textbf{AVG} ($\uparrow$) & \textbf{WGA} ($\uparrow$) & \textbf{AVG} ($\uparrow$) & \textbf{WGA} ($\uparrow$) & \textbf{AVG} ($\uparrow$) & \textbf{WGA} ($\uparrow$) \\
  \midrule
  \multicolumn{9}{c}{\textit{\textbf{Zero-Shot Evaluation (No Downstream Labels)}}} \\
  \midrule
  Native Backbone & $74.5 \std{0.0}$ & $51.3 \std{0.0}$ & $74.5 \std{0.0}$ & $51.3 \std{0.0}$ & $72.4 \std{0.0}$ & $50.9 \std{0.0}$ & $72.4 \std{0.0}$ & $50.9 \std{0.0}$ \\
  RoboShot & $76.3 \std{0.0}$ & $58.0 \std{0.0}$ & $76.3 \std{0.0}$ & $58.0 \std{0.0}$ & $74.7 \std{0.0}$ & $54.8 \std{0.0}$ & $74.7 \std{0.0}$ & $54.8 \std{0.0}$ \\
  DIAL & $78.4 \std{0.0}$ & $59.7 \std{0.0}$ & $78.4 \std{0.0}$ & $59.7 \std{0.0}$ & $77.9 \std{0.0}$ & $56.4 \std{0.0}$ & $77.9 \std{0.0}$ & $56.4 \std{0.0}$ \\
\textbf{BAP (ZS)} (Ours) & $\textbf{87.3} \std{0.0}$ & $\textbf{72.1} \std{0.0}$ & $\textbf{87.3} \std{0.0}$ & $\textbf{72.1} \std{0.0}$ & $\textbf{88.7} \std{0.0}$ & $\textbf{59.1} \std{0.0}$ & $\textbf{88.7} \std{0.0}$ & $\textbf{59.1} \std{0.0}$ \\
  \midrule
  \multicolumn{9}{c}{\textit{\textbf{Trained Evaluation (Linear Probing \& Fine-Tuning)}}} \\
  \midrule
  Native Backbone (LP) & $79.8 \std{0.6}$ & $61.6 \std{0.9}$ & $62.4 \std{0.3}$ & $24.5 \std{0.9}$ & $83.6 \std{0.7}$ & $67.4 \std{1.6}$ & $63.2 \std{0.8}$ & $21.9 \std{1.2}$ \\
  AFR & $91.5 \std{0.5}$ & $89.1 \std{1.6}$ & - & - & $92.1 \std{0.7}$ & $91.3 \std{1.2}$ & - & - \\
  DFR & $93.4 \std{0.5}$ & $91.3 \std{0.6}$ & - & - & $94.6 \std{0.3}$ & $92.2 \std{0.5}$ & - & - \\
  ERM (LP-FT) & $88.8 \std{0.2}$ & $72.6 \std{0.4}$ & $61.7 \std{0.7}$ & $23.8 \std{1.2}$ & $86.9 \std{0.8}$ & $70.3 \std{0.6}$ & $59.2 \std{1.1}$ & $19.4 \std{1.6}$ \\
  WiSE-FT & $87.5 \std{0.1}$ & $67.0 \std{0.3}$ & $61.8 \std{0.4}$ & $22.0 \std{0.3}$ & $85.1 \std{0.5}$ & $67.3 \std{0.4}$ & $55.9 \std{0.8}$ & $21.7 \std{1.1}$ \\
  PruSC & $77.3 \std{2.8}$ & $56.4 \std{3.9}$ & $46.5 \std{3.1}$ & $36.9 \std{4.2}$ & $80.3 \std{2.1}$ & $57.1 \std{2.9}$ & $49.1 \std{2.3}$ & $38.8 \std{3.7}$ \\
  Data-Matched Control & $91.1 \std{1.4}$ & $82.7 \std{3.6}$ & $86.7 \std{2.7}$ & $71.4 \std{6.9}$ & $93.8 \std{0.6}$ & $88.3 \std{1.1}$ & $88.9 \std{1.0}$ & $76.8 \std{1.2}$ \\
  \textbf{BAP (LP)} (Ours) & $\mathbf{94.3 \std{0.6}}$ & $\mathbf{92.1 \std{1.2}}$ & $\mathbf{94.2 \std{0.5}}$ & $\mathbf{91.8 \std{0.9}}$ & $\mathbf{95.3 \std{0.7}}$ & $\mathbf{93.2 \std{0.9}}$ & $\mathbf{95.6 \std{0.7}}$ & $\mathbf{92.7 \std{0.8}}$ \\
  \bottomrule
  \end{tabular}
  \end{small}
\end{table}

The results in Table \ref{tab:waterbirds} demonstrate that BAP consistently outperforms established spurious correlation mitigation techniques for Waterbirds-$95\%$. Notably, BAP slightly exceeds the performance of the DFR oracle on Waterbirds-$95\%$ while offering higher performance than the data-matched control indicating that the robustness induced by our method is not exclusively due to increased data exposure. The gap in WGA grows when investigating the results for Waterbirds-$100\%$. BAP maintains a WGA above $90\%$ despite a $100\%$ spurious correlation without reliance on minority-group annotations (to our knowledge, a new state-of-the-art). Furthermore, BAP exhibits nearly identical performance across both $95\%$ and $100\%$ correlated datasets suggesting our pre-training regime facilitates robust background invariance regardless of the strength of spurious correlations in the downstream dataset. The ability to deploy BAP pre-trained models on datasets lacking minority groups without performance degradation offers unique practical advantages over existing methods.

We observe a similar pattern with the gender classification task in Table \ref{tab:gender} where BAP maintains near identical performance across both $95\%$ and $100\%$ regimes in CLIP and SigLIP 2. While the data matched control baseline outperforms BAP in the $95\%$ regime using SigLIP 2, its performance declines in the $100\%$ regime indicating that data exposure may match BAP's performance in certain contexts; however, BAP remains the only baseline with no noticeable robustness decline when going from $95\%$ to $100\%$ regimes (another state-of-the-art to our knowledge). The results in Tables \ref{tab:waterbirds} and \ref{tab:gender} demonstrate that robustness requires more than mere data exposure; BAP successfully outperforms the data-matched control despite the latter's explicit advantage of supervised training on ground-truth labels.

\begin{table}[h]
  \caption{Average and worst-group accuracies (mean $\pm$ std over 5 runs) across various baselines for the COCO gender classification benchmark using 95\% and 100\%  spurious correlation rates.}
  \label{tab:gender}
  \centering
  \setlength{\tabcolsep}{2.5pt} 
  \begin{small} 
  \begin{tabular}{l cccc cccc}
  \toprule
  & \multicolumn{4}{c}{\textbf{CLIP}} & \multicolumn{4}{c}{\textbf{SigLIP2}} \\
  \cmidrule(lr){2-5} \cmidrule(lr){6-9}
  & \multicolumn{2}{c}{\textbf{COCO-95\%}} & \multicolumn{2}{c}{\textbf{COCO-100\%}} & \multicolumn{2}{c}{\textbf{COCO-95\%}} & \multicolumn{2}{c}{\textbf{COCO-100\%}} \\
  \cmidrule(lr){2-3} \cmidrule(lr){4-5} \cmidrule(lr){6-7} \cmidrule(lr){8-9}
  \textbf{Method} & \textbf{AVG} ($\uparrow$) & \textbf{WGA} ($\uparrow$) & \textbf{AVG} ($\uparrow$) & \textbf{WGA} ($\uparrow$) & \textbf{AVG} ($\uparrow$) & \textbf{WGA} ($\uparrow$) & \textbf{AVG} ($\uparrow$) & \textbf{WGA} ($\uparrow$) \\
  \midrule
  \multicolumn{9}{c}{\textit{\textbf{Zero-Shot Evaluation (No Downstream Labels)}}} \\
  \midrule
  Native Backbone & $71.4 \std{0.0}$ & $48.3 \std{0.0}$ & $71.4 \std{0.0}$ & $48.3 \std{0.0}$ & $71.3 \std{0.0}$ & $64.2 \std{0.0}$ & $71.3 \std{0.0}$ & $64.2 \std{0.0}$ \\
  RoboShot & $73.2 \std{0.0}$ & $52.5 \std{0.0}$ & $73.2 \std{0.0}$ & $52.5 \std{0.0}$ & $72.9 \std{0.0}$ & $67.0 \std{0.0}$ & $72.9 \std{0.0}$ & $67.0 \std{0.0}$ \\
  DIAL & $\textbf{75.1} \std{0.0}$ & $53.8 \std{0.0}$ & $ \textbf{75.1} \std{0.0}$ & $53.8 \std{0.0}$ & $72.4 \std{0.0}$ & $65.2 \std{0.0}$ & $72.4 \std{0.0}$ & $65.2 \std{0.0}$ \\
  \textbf{BAP (ZS)} (Ours) & $74.2 \std{0.0}$ & $\textbf{57.1} \std{0.0}$ & $74.2 \std{0.0}$ & $\textbf{57.1} \std{0.0}$ & $\textbf{77.5} \std{0.0}$ & $\textbf{71.8} \std{0.0}$ & $\textbf{77.5} \std{0.0}$ & $\textbf{71.8} \std{0.0}$ \\
  \midrule
  \multicolumn{9}{c}{\textit{\textbf{Trained Evaluation (Linear Probing \& Fine-Tuning)}}} \\
  \midrule
  Native Backbone (LP) & $57.3 \std{0.6}$ & $23.4 \std{0.8}$ & $49.6 \std{0.7}$ & $5.3 \std{0.2}$ & $58.9 \std{0.6}$ & $32.3 \std{0.8}$ & $51.4 \std{0.8}$ & $9.3 \std{1.1}$ \\
  AFR & $73.3 \std{1.4}$ & $64.7 \std{1.7}$ & - & - & $56.8 \std{1.1}$ & $68.6 \std{1.4}$ & - & - \\
  DFR & $74.2 \std{0.9}$ & $66.8 \std{1.1}$ & - & - & $75.2 \std{0.9}$ & $71.4 \std{1.1}$ & - & - \\
  ERM (LP-FT) & $60.3 \std{0.7}$ & $54.2 \std{0.9}$ & $63.1 \std{1.1}$ & $4.9 \std{0.5}$ & $63.3 \std{0.5}$ & $41.2 \std{0.7}$ & $55.3 \std{0.6}$ & $6.2 \std{0.9}$ \\
  WiSE-FT & $63.1 \std{0.4}$ & $57.4 \std{0.6}$ & $65.5 \std{0.8}$ & $32.7 \std{1.1}$ & $65.7 \std{0.6}$ & $53.2 \std{0.8}$ & $54.9 \std{0.8}$ & $36.1 \std{1.3}$ \\
  PruSC & $55.6 \std{2.2}$ & $48.6 \std{2.7}$ & $53.0 \std{2.4}$ & $11.5 \std{1.9}$ & $57.5 \std{2.8}$ & $45.9 \std{3.6}$ & $54.6 \std{3.1}$ & $13.7 \std{2.8}$ \\
  Data-Matched Control & $75.2 \std{0.4}$ & $73.8 \std{0.7}$ & $73.2 \std{0.8}$ & $64.6 \std{1.1}$ & $\mathbf{78.1} \std{1.1}$ & $\mathbf{76.2} \std{1.3}$ & $74.8 \std{0.7}$ & $63.5 \std{0.9}$ \\
  \textbf{BAP (LP)} (Ours) & $\mathbf{78.1 \std{1.1}}$ & $\mathbf{75.7 \std{1.2}}$ & $\mathbf{77.9 \std{1.0}}$ & $\mathbf{74.7 \std{1.2}}$ & $77.6 \std{0.8}$ & $75.8 \std{0.9}$ & $\mathbf{77.8} \std{0.9}$ & $\mathbf{75.3} \std{1.2}$ \\
  \bottomrule
  \end{tabular}
  \end{small}
\end{table}

\subsection{ Real-world evaluation: CounterAnimal and NICO++}\label{sec: counteranimal}

Results from Section \ref{sec: waterbirds results}, while promising, do not establish whether robustness gained from synthetic composites carries over to real images. To test the performance of BAP in real world settings, we adapt our bird classification task to the CounterAnimal \citep{sober_clip} benchmark which consists of natural images containing animals in common contexts (e.g., polar bears on ice) and counter-contexts (e.g., polar bears on grass) which allows us to investigate BAPs sim-to-real performance.

We enforce a $100\%$ spurious correlation which presents a dual challenge: the model is required to overcome both a perfect spurious correlation and a strong sim-to-real distribution shift. We evaluate the robustness of BAP across two distinct downstream binary classification tasks with moderately challenging pairings, as the bird pairs share taxonomic orders, silhouettes, and camouflage. 

\begin{table}[ht]
\centering
\caption{Average and worst-group accuracies (mean $\pm$ std over 10 runs) across various baselines for two pairs of binary classification on the CounterAnimal dataset. (\textbf{Top}) Zero shot evaluations; \textbf{Bottom} Trained evaluations.}
\vspace{-1pt}
\label{tab:counteranimal results}
\vskip 0.15in
\small
\setlength{\tabcolsep}{2.5pt} 
\begin{tabular}{l cccc cccc} 
\toprule
& \multicolumn{4}{c}{\textbf{Pair 1: Brambling vs. Bulbul}} & \multicolumn{4}{c}{\textbf{Pair 2: Ptarmigan vs. Prairie-Chicken}} \\
\cmidrule(lr){2-5} \cmidrule(lr){6-9}
& \multicolumn{2}{c}{CLIP} & \multicolumn{2}{c}{SigLIP2} & \multicolumn{2}{c}{CLIP} & \multicolumn{2}{c}{SigLIP2} \\
\cmidrule(lr){2-3} \cmidrule(lr){4-5} \cmidrule(lr){6-7} \cmidrule(lr){8-9}
\textbf{Method} & \textbf{AVG} ($\uparrow$) & \textbf{WGA} ($\uparrow$) & \textbf{AVG} ($\uparrow$) & \textbf{WGA} ($\uparrow$) & \textbf{AVG} ($\uparrow$) & \textbf{WGA} ($\uparrow$) & \textbf{AVG} ($\uparrow$) & \textbf{WGA} ($\uparrow$) \\
\midrule

Native Backbone (ZS)                 & 78.5 \std{0.0}          & 56.3 \std{0.0}          & $70.4 \std{0.0}$ & $36.3 \std{0.0}$ & 81.6 \std{0.0}          & 46.4 \std{0.0}          & $81.4 \std{0.0}$ & $44.7 \std{0.0}$ \\
DIAL                   & 79.2 \std{0.0}          & 61.6 \std{0.0}          & $74.6 \std{0.0}$ & $42.0 \std{0.0}$ & 76.7 \std{0.0}          & 51.3 \std{0.0}          & $\textbf{84.1} \std{0.0}$ & $53.9 \std{0.0}$ \\

\textbf{BAP (ZS)} (Ours) & $\textbf{83.5} \std{0.0}$ & $\textbf{64.9} \std{0.0}$ & $\textbf{81.9} \std{0.0}$ & $\textbf{56.7} \std{0.0}$ & $\textbf{79.1} \std{0.0}$ & $\textbf{55.8} \std{0.0}$ & $83.8 \std{0.0}$ & $\textbf{59.2} \std{0.0}$ \\

\midrule
\midrule

Native Backbone (LP)               & 91.7 \std{0.6}          & 74.2 \std{3.1}          & $92.4 \std{0.4}$ & $71.8 \std{2.6}$ & 80.5 \std{1.4}          & 50.4 \std{8.2}          & $70.9 \std{1.0}$ & $46.3 \std{6.3}$ \\
ERM (LP-FT)                & 68.6 \std{4.3}          & 23.2 \std{8.6}          & $73.1 \std{3.1}$ & $31.9 \std{5.1}$ & 65.3 \std{6.4}          & 19.2 \std{4.9}          & $61.8 \std{4.4}$ & $14.7 \std{5.6}$ \\
PRuSC                      & 52.1 \std{3.7}          & 34.6 \std{7.2}          & $57.9 \std{2.9}$ & $41.3 \std{5.2}$ & 47.1 \std{4.7}          & 29.7 \std{5.2}          & $52.8 \std{6.2}$ & $ 21.6\std{7.9}$ \\
Data-Matched Control       & 86.8 \std{5.6}          & 70.4 \std{6.3}         & $85.2 \std{3.6}$ & $73.6 \std{5.7}$ & 67.0 \std{4.4}          & 43.9 \std{5.1}          & $75.7 \std{2.9}$ & $61.3 \std{6.1}$ \\

\textbf{BAP (LP)} (Ours)          & \textbf{92.0 \std{0.7}} & \textbf{86.4 \std{3.3}} & $\textbf{93.2} \std{0.8}$ & $\textbf{88.3} \std{2.5}$ & \textbf{81.4 \std{2.2}} & \textbf{65.0 \std{3.7}} & $\textbf{80.4} \std{1.7}$ & $\textbf{69.3} \std{3.2}$ \\

\bottomrule
\end{tabular}
\end{table}

The results in Table \ref{tab:counteranimal results} demonstrate that the robustness gains facilitated by BAP successfully transfer to natural images. In both evaluations, BAP-pre-trained encoders yield an improvement of over $10$ percentage points in WGA compared to native pre-trained encoders. Furthermore, we observe a significant performance gap between BAP and the Data-Matched ERM Control suggesting our anchor extraction and alignment protocol facilitates a degree of background invariance that cannot be attributed to data exposure alone. This consistent performance across both synthetic Waterbirds and real CounterAnimal benchmarks highlights BAP's utility as a robust feature extractor for practical deployment.

\begin{table}[ht]
\centering
\caption{Average and worst-group accuracies (mean $\pm$ std over 5 runs) across the best performing baselines for four pairs of binary vehicle classification on the NICO++ dataset using CLIP ConvNeXt-W.  (Note: results obtained using SigLIP2 can be seen in Table \ref{tab:nico_vit}).}
\label{tab:nico++ results}
\small
\setlength{\tabcolsep}{2.3pt}
\begin{tabular}{l cccccccc}
\toprule
& \multicolumn{2}{c}{\textbf{Car vs. Truck}} 
& \multicolumn{2}{c}{\textbf{Car vs. Bus}} 
& \multicolumn{2}{c}{\textbf{Ship vs. Sailboat}} 
& \multicolumn{2}{c}{\textbf{Bike vs. Motorbike}} \\
\cmidrule(lr){2-3} \cmidrule(lr){4-5} \cmidrule(lr){6-7} \cmidrule(lr){8-9}
\textbf{Method} 
& \textbf{AVG} ($\uparrow$) & \textbf{WGA} ($\uparrow$) 
& \textbf{AVG} ($\uparrow$) & \textbf{WGA} ($\uparrow$) 
& \textbf{AVG} ($\uparrow$) & \textbf{WGA} ($\uparrow$)
& \textbf{AVG} ($\uparrow$) & \textbf{WGA} ($\uparrow$) \\
\midrule
Native Backbone (ZS)
& 79.0 \std{0.0} & 56.5 \std{0.0} 
& \textbf{82.8} \std{0.0} & 58.2 \std{0.0} 
& 79.4 \std{0.0} & 53.7 \std{0.0}
& 75.8 \std{0.0} & 60.1 \std{0.0} \\

DIAL 
& 76.2 \std{0.0} & 58.3 \std{0.0} 
& 77.3 \std{0.0} & 64.2 \std{0.0} 
& 78.2 \std{0.0} & 61.5 \std{0.0}
& 73.9 \std{0.0} & 64.0 \std{0.0} \\

\textbf{BAP (ZS)} (Ours) 
& \textbf{79.0} \std{0.0} & \textbf{63.1} \std{0.0} 
& 81.6 \std{0.0} & \textbf{74.5} \std{0.0} 
& \textbf{82.4} \std{0.0} & \textbf{66.8} \std{0.0}
& \textbf{76.7} \std{0.0} & \textbf{69.8} \std{0.0} \\

\midrule
\midrule

Native Backbone (LP) 
& 79.6 \std{1.4} & 40.0 \std{3.1} 
& 81.1 \std{0.3} & 43.9 \std{0.8} 
& 79.9 \std{0.3} & 69.3 \std{1.4}
& 83.1 \std{0.9} & 55.3 \std{1.6} \\

Data-Matched Control 
& 81.8 \std{0.6} & 59.3 \std{2.8} 
& 84.5 \std{0.3} & 63.5 \std{0.4} 
& 82.4 \std{0.8} & 74.8 \std{1.3}
& \textbf{87.2} \std{1.7} & 70.1 \std{2.2} \\

\textbf{BAP (LP)} (Ours) 
& \textbf{83.8} \std{1.4} & \textbf{72.9} \std{2.0} 
& \textbf{87.0} \std{1.6} & \textbf{81.6} \std{2.3} 
& \textbf{84.6} \std{1.6} & \textbf{79.1} \std{2.7}
& 85.9 \std{2.1} & \textbf{80.3} \std{3.6} \\

\bottomrule
\end{tabular}
\end{table}

Finally, the NICO++ vehicle classification results are presented in Table \ref{tab:nico++ results}, alongside the most competitive baselines. BAP maintains a dominant lead across all tasks, achieving double-digit margins in WGA for the majority of results. We draw particular attention to the ``Car vs. Truck'' task, where BAP yields a \textbf{30-percentage-point increase in WGA} over the native CLIP representations. A similar pattern is visible in the ``Bike vs. Motorbike'' task. Since `truck' and `motorbike' instances were explicitly omitted from the synthetic pre-training object pool, this result strongly indicates that BAP successfully recalibrates the model's inductive bias toward super-class-wide background invariance, rather than relying on category-specific memorization.

\section{Discussion, limitations, and practical aspects}\label{sec:conclusion}

BAP successfully leverages linear additivity \citep{linear_CLIP, linear_VLM1, linear_VLM2} to establish a robust, task-agnostic pre-training step capable of withstanding $100\%$ spurious correlations. By securing state-of-the-art worst-group accuracy even in the complete absence of minority-group data, it offers a highly practical solution for real-world deployment. Nevertheless, this specialized focus on background invariance introduces distinct trade-offs. Firstly, we acknowledge that BAP is more limited in scope than general strategies like DFR or DIAL, as it is specifically formulated for spatially separable correlations and would not function where the spurious feature is part of the core object, such as in CelebA \citep{celebA}. However, given the robustness profile exhibited by BAP and the ease of downstream deployment, we believe our method is superior for the specific case of background spurious correlations. Furthermore, it is well established in the robustness literature that enforcing strict invariance or high robustness incurs a cost to generalizability \citep{robust_tradeoff}. BAP intentionally sacrifices the backbone's open-world background recognition capabilities to achieve state-of-the-art OOD foreground robustness. This trade-off is highly desirable in high-stakes domains like medical imaging \citep{medical_one,medical_two} and is investigated fully in Appendix \ref{app: bap zero shot}.

Finally, we note that BAP relies on segmented foreground items, which may be difficult to obtain. To assess practicality, we conduct two ablations on CLIP evaluated on Waterbirds-$100\%$ (SigLIP 2 results and full methodology can be seen in Appendix \ref{app:practical}). First, we perform a sensitivity analysis of segmentation quality across four settings: \textbf{Perfect} masks, dilated masks introducing background noise (\textbf{Noisy}), eroded masks removing object details (\textbf{Botched}), and coarse \textbf{Bounding Box}es—spanning a spectrum from ideal isolation to increasingly noisy or incomplete foregrounds. Results in Table \ref{tab:seg_ablation} show that BAP is robust to imperfect masks and may offer robustness gains even with coarse bounding boxes.

\begin{figure}[ht]
    \centering
    
    \begin{minipage}[t]{0.52\textwidth}
        \vspace{0pt} 
        \centering
        \captionof{table}{Sensitivity Analysis of Segmentation Quality across Waterbirds-100\%. We report AVG and WGA across degradation modes (mean $\pm$ std over 5 runs). BAP maintains robustness even with imperfect segmentations (Noisy and Botched) while even providing notable robustness gains using bounding boxes in place of pixel-level segmentation masks.}
        \label{tab:seg_conc} 
        \setlength{\tabcolsep}{6pt}
        \vspace{5pt}
\begin{tabular}{l cc}
        \toprule
        \textbf{Segmentation Quality} & \textbf{AVG} & \textbf{WGA} \\ 
        \midrule
\textbf{Perfect}      & $94.1 \std{0.6}$ & $91.6 \std{0.8}$ \\
\textbf{Noisy}        & $93.3 \std{0.5}$ & $89.9 \std{1.7}$ \\
\textbf{Botched}      & $92.6 \std{0.6}$ & $90.5 \std{1.4}$ \\
\textbf{Bounding Box} & $89.8 \std{1.5}$ & $82.5 \std{2.3}$ \\
        \bottomrule
\end{tabular}
    \end{minipage}
    \hfill 
    \begin{minipage}[t]{0.42\textwidth}
        \vspace{0pt} 
        \centering
        \includegraphics[trim={0cm} {0.55cm} {0.1cm} {0.1cm}, clip,width=\linewidth]{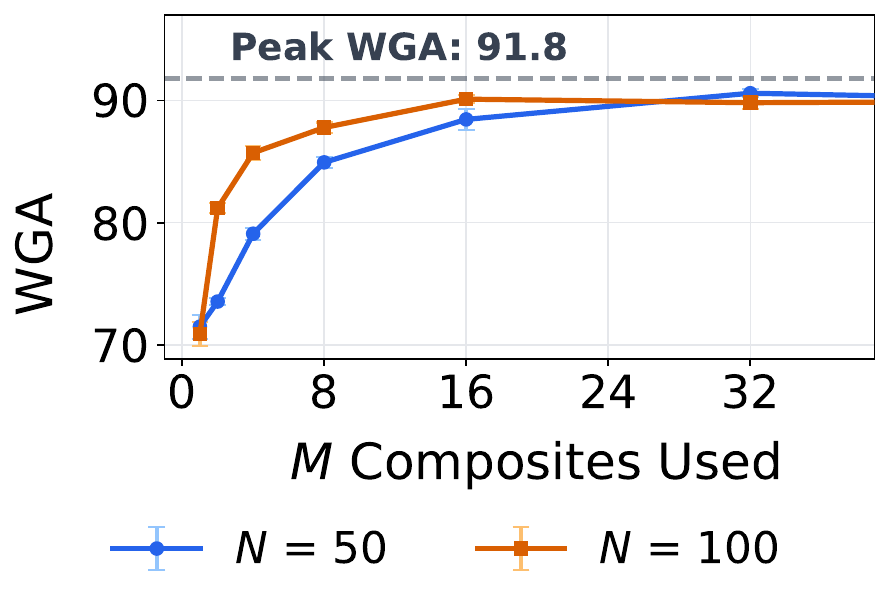}
        \captionof{figure}{Effect of parameter $M$ on BAP using CLIP at low values of $N$ showing that high performance may be achieved with as few as $50$ distinct segmented items.}
        \label{fig:m_conc}
    \end{minipage}
    
\end{figure}

Second, we simulate data-scarce conditions by running BAP with only $N = 50$ and $N = 100$ distinct foreground items. We offset the reduced $N$ by increasing the number of random contexts ($M$) each item appears in during Phase 2 (distinct from the number of composites $K$ used for anchor construction). Results in Figure \ref{fig:m_conc} show that while performance drops at low $M$ ($<8$), it recovers at higher $M$ ($\ge 16$), remaining within $3\%$ of the full dataset ($N \approx 5{,}000$). Thus, increasing $M$ can recover near-peak robustness under limited foreground data.

The results of the ablations in Table \ref{tab:seg_conc} and Figure \ref{fig:m_conc} indicate that BAP is highly practical and that the requirement for segmentation masks is a manageable pre-requisite which may easily be satisfied with automated tools such as the Segment Anything Model \citep{SAM}. We believe that targeted, representation level interventions such as BAP may replace existing general post-hoc retraining methods and that BAP may serve to as an exemplar for such approaches.

\newpage
\bibliographystyle{plainnat}
\bibliography{example_paper}

\begin{thebibliography}{47}
\providecommand{\natexlab}[1]{#1}
\providecommand{\url}[1]{\texttt{#1}}
\expandafter\ifx\csname urlstyle\endcsname\relax
  \providecommand{\doi}[1]{doi: #1}\else
  \providecommand{\doi}{doi: \begingroup \urlstyle{rm}\Url}\fi

\bibitem[Adila et~al.(2023)Adila, Shin, Cai, and Sala]{roboshot}
Dyah Adila, Changho Shin, Linrong Cai, and Frederic Sala.
\newblock Zero-shot robustification of zero-shot models.
\newblock \emph{arXiv preprint arXiv:2309.04344}, 2023.

\bibitem[Agarwal et~al.(2025)Agarwal, Karanam, and Gandhi]{foreground_or_background}
Aishwarya Agarwal, Srikrishna Karanam, and Vineet Gandhi.
\newblock Foreground or background? visual interpretability and robustness analysis of {CLIP}, 2025.
\newblock URL \url{https://openreview.net/forum?id=K7wkjqLjrt}.

\bibitem[Bhalla et~al.(2024)Bhalla, Oesterling, Srinivas, Calmon, and Lakkaraju]{linear_CLIP}
Usha Bhalla, Alex Oesterling, Suraj Srinivas, Flavio~P Calmon, and Himabindu Lakkaraju.
\newblock Interpreting clip with sparse linear concept embeddings (splice).
\newblock \emph{Advances in Neural Information Processing Systems}, 37:\penalty0 84298--84328, 2024.

\bibitem[Csurka et~al.(2004)Csurka, Dance, Fan, Willamowski, and Bray]{bag}
Gabriella Csurka, Christopher Dance, Lixin Fan, Jutta Willamowski, and C{\'e}dric Bray.
\newblock Visual categorization with bags of keypoints.
\newblock In \emph{Workshop on statistical learning in computer vision, ECCV}, volume~1, pages 1--2. Prague, 2004.

\bibitem[Dehkharghanian et~al.(2023)Dehkharghanian, Bidgoli, Riasatian, Mazaheri, Campbell, Pantanowitz, Tizhoosh, and Rahnamayan]{medical_two}
Taher Dehkharghanian, Azam~Asilian Bidgoli, Abtin Riasatian, Pooria Mazaheri, Clinton~JV Campbell, Liron Pantanowitz, HR~Tizhoosh, and Shahryar Rahnamayan.
\newblock Biased data, biased ai: deep networks predict the acquisition site of tcga images.
\newblock \emph{Diagnostic Pathology}, 18\penalty0 (1):\penalty0 67, 2023.

\bibitem[Deng et~al.(2009)Deng, Dong, Socher, Li, Li, and Fei-Fei]{IN1K}
Jia Deng, Wei Dong, Richard Socher, Li-Jia Li, Kai Li, and Li~Fei-Fei.
\newblock Imagenet: A large-scale hierarchical image database.
\newblock In \emph{2009 IEEE conference on computer vision and pattern recognition}, pages 248--255. Ieee, 2009.

\bibitem[Devlin et~al.(2019)Devlin, Chang, Lee, and Toutanova]{BERT}
Jacob Devlin, Ming-Wei Chang, Kenton Lee, and Kristina Toutanova.
\newblock Bert: Pre-training of deep bidirectional transformers for language understanding, 2019.
\newblock URL \url{https://arxiv.org/abs/1810.04805}.

\bibitem[Dosovitskiy et~al.(2021)Dosovitskiy, Beyer, Kolesnikov, Weissenborn, Zhai, Unterthiner, Dehghani, Minderer, Heigold, Gelly, Uszkoreit, and Houlsby]{vit}
Alexey Dosovitskiy, Lucas Beyer, Alexander Kolesnikov, Dirk Weissenborn, Xiaohua Zhai, Thomas Unterthiner, Mostafa Dehghani, Matthias Minderer, Georg Heigold, Sylvain Gelly, Jakob Uszkoreit, and Neil Houlsby.
\newblock An image is worth 16x16 words: Transformers for image recognition at scale, 2021.
\newblock URL \url{https://arxiv.org/abs/2010.11929}.

\bibitem[Gadre et~al.(2023)Gadre, Ilharco, Fang, Hayase, Smyrnis, Nguyen, Marten, Wortsman, Ghosh, Zhang, et~al.]{datacomp}
Samir~Yitzhak Gadre, Gabriel Ilharco, Alex Fang, Jonathan Hayase, Georgios Smyrnis, Thao Nguyen, Ryan Marten, Mitchell Wortsman, Dhruba Ghosh, Jieyu Zhang, et~al.
\newblock Datacomp: In search of the next generation of multimodal datasets.
\newblock \emph{Advances in Neural Information Processing Systems}, 36:\penalty0 27092--27112, 2023.

\bibitem[Gao et~al.(2024)Gao, Geng, Zhang, Ma, Fang, Zhang, Li, and Qiao]{clip_adapt}
Peng Gao, Shijie Geng, Renrui Zhang, Teli Ma, Rongyao Fang, Yongfeng Zhang, Hongsheng Li, and Yu~Qiao.
\newblock Clip-adapter: Better vision-language models with feature adapters.
\newblock \emph{International Journal of Computer Vision}, 132\penalty0 (2):\penalty0 581--595, 2024.

\bibitem[He et~al.(2022)He, Chen, Xie, Li, Doll{\'a}r, and Girshick]{MAE}
Kaiming He, Xinlei Chen, Saining Xie, Yanghao Li, Piotr Doll{\'a}r, and Ross Girshick.
\newblock Masked autoencoders are scalable vision learners.
\newblock In \emph{Proceedings of the IEEE/CVF conference on computer vision and pattern recognition}, pages 16000--16009, 2022.

\bibitem[Higgins et~al.(2017{\natexlab{a}})Higgins, Matthey, Pal, Burgess, Glorot, Botvinick, Mohamed, and Lerchner]{beta-VAE}
Irina Higgins, Loic Matthey, Arka Pal, Christopher Burgess, Xavier Glorot, Matthew Botvinick, Shakir Mohamed, and Alexander Lerchner.
\newblock beta-{VAE}: Learning basic visual concepts with a constrained variational framework.
\newblock In \emph{International Conference on Learning Representations}, 2017{\natexlab{a}}.
\newblock URL \url{https://openreview.net/forum?id=Sy2fzU9gl}.

\bibitem[Higgins et~al.(2017{\natexlab{b}})Higgins, Matthey, Pal, Burgess, Glorot, Botvinick, Mohamed, and Lerchner]{bvae}
Irina Higgins, Loic Matthey, Arka Pal, Christopher Burgess, Xavier Glorot, Matthew Botvinick, Shakir Mohamed, and Alexander Lerchner.
\newblock beta-{VAE}: Learning basic visual concepts with a constrained variational framework.
\newblock In \emph{International Conference on Learning Representations}, 2017{\natexlab{b}}.
\newblock URL \url{https://openreview.net/forum?id=Sy2fzU9gl}.

\bibitem[Janouskova et~al.(2025)Janouskova, Gavrus, and Matas]{janouskova2025robust}
Klara Janouskova, Cristian Gavrus, and Jiri Matas.
\newblock Robust context-aware object recognition.
\newblock \emph{arXiv preprint arXiv:2510.00618}, 2025.

\bibitem[Khan et~al.(2022)Khan, Naseer, Hayat, Zamir, Khan, and Shah]{transformer_survey}
Salman Khan, Muzammal Naseer, Munawar Hayat, Syed~Waqas Zamir, Fahad~Shahbaz Khan, and Mubarak Shah.
\newblock Transformers in vision: A survey.
\newblock \emph{ACM Computing Surveys (CSUR)}, 54\penalty0 (10s):\penalty0 1--41, 2022.

\bibitem[Kirichenko et~al.(2022)Kirichenko, Izmailov, and Wilson]{dfr}
Polina Kirichenko, Pavel Izmailov, and Andrew~Gordon Wilson.
\newblock Last layer re-training is sufficient for robustness to spurious correlations.
\newblock \emph{arXiv preprint arXiv:2204.02937}, 2022.

\bibitem[Kirillov et~al.(2023)Kirillov, Mintun, Ravi, Mao, Rolland, Gustafson, Xiao, Whitehead, Berg, Lo, et~al.]{SAM}
Alexander Kirillov, Eric Mintun, Nikhila Ravi, Hanzi Mao, Chloe Rolland, Laura Gustafson, Tete Xiao, Spencer Whitehead, Alexander~C Berg, Wan-Yen Lo, et~al.
\newblock Segment anything.
\newblock In \emph{Proceedings of the IEEE/CVF international conference on computer vision}, pages 4015--4026, 2023.

\bibitem[Kumar et~al.(2022)Kumar, Raghunathan, Jones, Ma, and Liang]{lp_then_ft}
Ananya Kumar, Aditi Raghunathan, Robbie Jones, Tengyu Ma, and Percy Liang.
\newblock Fine-tuning can distort pretrained features and underperform out-of-distribution.
\newblock \emph{arXiv preprint arXiv:2202.10054}, 2022.

\bibitem[Le et~al.(2024)Le, Schl{\"o}tterer, and Seifert]{prusc}
Phuong~Quynh Le, J{\"o}rg Schl{\"o}tterer, and Christin Seifert.
\newblock Out of spuriousity: Improving robustness to spurious correlations without group annotations.
\newblock \emph{arXiv preprint arXiv:2407.14974}, 2024.

\bibitem[Lin et~al.(2014)Lin, Maire, Belongie, Hays, Perona, Ramanan, Doll{\'a}r, and Zitnick]{coco}
Tsung-Yi Lin, Michael Maire, Serge Belongie, James Hays, Pietro Perona, Deva Ramanan, Piotr Doll{\'a}r, and C~Lawrence Zitnick.
\newblock Microsoft coco: Common objects in context.
\newblock In \emph{European Conference on Computer Vision}, pages 740--755. Springer, 2014.

\bibitem[Liu et~al.(2025)Liu, Liu, Lai, Shen, Zhao, and Lei]{waterbirds100}
Chenruo Liu, Hongjun Liu, Zeyu Lai, Yiqiu Shen, Chen Zhao, and Qi~Lei.
\newblock Superclass-guided representation disentanglement for spurious correlation mitigation, 2025.
\newblock URL \url{https://arxiv.org/abs/2508.08570}.

\bibitem[Liu et~al.(2022)Liu, Mao, Wu, Feichtenhofer, Darrell, and Xie]{convnext}
Zhuang Liu, Hanzi Mao, Chao-Yuan Wu, Christoph Feichtenhofer, Trevor Darrell, and Saining Xie.
\newblock A convnet for the 2020s, 2022.
\newblock URL \url{https://arxiv.org/abs/2201.03545}.

\bibitem[Liu et~al.(2015)Liu, Luo, Wang, and Tang]{celebA}
Ziwei Liu, Ping Luo, Xiaogang Wang, and Xiaoou Tang.
\newblock Deep learning face attributes in the wild, 2015.
\newblock URL \url{https://arxiv.org/abs/1411.7766}.

\bibitem[Maheronnaghsh and Alvanagh(2024)]{clip_bias_1}
Mohammadjavad Maheronnaghsh and Taha~Akbari Alvanagh.
\newblock Robustness to spurious correlation: A comprehensive review.
\newblock In \emph{European Conference on Computer Vision}, pages 361--379. Springer, 2024.

\bibitem[Mikolov et~al.(2013)Mikolov, Chen, Corrado, and Dean]{word2vec}
Tomas Mikolov, Kai Chen, Greg Corrado, and Jeffrey Dean.
\newblock Efficient estimation of word representations in vector space, 2013.
\newblock URL \url{https://arxiv.org/abs/1301.3781}.

\bibitem[Oquab et~al.(2023)Oquab, Darcet, Moutakanni, Vo, Szafraniec, Khalidov, Fernandez, Haziza, Massa, El-Nouby, et~al.]{DinoV2}
Maxime Oquab, Timoth{\'e}e Darcet, Th{\'e}o Moutakanni, Huy Vo, Marc Szafraniec, Vasil Khalidov, Pierre Fernandez, Daniel Haziza, Francisco Massa, Alaaeldin El-Nouby, et~al.
\newblock Dinov2: Learning robust visual features without supervision.
\newblock \emph{arXiv preprint arXiv:2304.07193}, 2023.

\bibitem[Paduraru et~al.()Paduraru, Barbalau, Filipescu, Nicolicioiu, and Burceanu]{BEE}
Cristian~Daniel Paduraru, Antonio Barbalau, Radu Filipescu, Andrei~Liviu Nicolicioiu, and Elena Burceanu.
\newblock Bridging explainability and embeddings: Bee aware of spuriousness.
\newblock In \emph{The Fourteenth International Conference on Learning Representations}.

\bibitem[Papadimitriou et~al.(2025)Papadimitriou, Su, Fel, Kakade, and Gil]{linear_VLM2}
Isabel Papadimitriou, Huangyuan Su, Thomas Fel, Sham Kakade, and Stephanie Gil.
\newblock Interpreting the linear structure of vision-language model embedding spaces.
\newblock \emph{arXiv preprint arXiv:2504.11695}, 2025.

\bibitem[Qiu et~al.(2023)Qiu, Potapczynski, Izmailov, and Wilson]{afr}
Shikai Qiu, Andres Potapczynski, Pavel Izmailov, and Andrew~Gordon Wilson.
\newblock Simple and fast group robustness by automatic feature reweighting, 2023.
\newblock URL \url{https://arxiv.org/abs/2306.11074}.

\bibitem[Radford et~al.(2021)Radford, Kim, Hallacy, Ramesh, Goh, Agarwal, Sastry, Askell, Mishkin, Clark, et~al.]{CLIP_original}
Alec Radford, Jong~Wook Kim, Chris Hallacy, Aditya Ramesh, Gabriel Goh, Sandhini Agarwal, Girish Sastry, Amanda Askell, Pamela Mishkin, Jack Clark, et~al.
\newblock Learning transferable visual models from natural language supervision.
\newblock In \emph{International Conference on Machine Learning}, pages 8748--8763. PmLR, 2021.

\bibitem[Ridnik et~al.(2021)Ridnik, Ben-Baruch, Noy, and Zelnik-Manor]{IN21k}
Tal Ridnik, Emanuel Ben-Baruch, Asaf Noy, and Lihi Zelnik-Manor.
\newblock Imagenet-21k pretraining for the masses.
\newblock \emph{arXiv preprint arXiv:2104.10972}, 2021.

\bibitem[Sagawa et~al.(2020)Sagawa, Koh, Hashimoto, and Liang]{waterbirds}
Shiori Sagawa, Pang~Wei Koh, Tatsunori~B. Hashimoto, and Percy Liang.
\newblock Distributionally robust neural networks for group shifts: On the importance of regularization for worst-case generalization, 2020.
\newblock URL \url{https://arxiv.org/abs/1911.08731}.

\bibitem[Schuhmann et~al.(2022)Schuhmann, Beaumont, Vencu, Gordon, Wightman, Cherti, Coombes, Katta, Mullis, Wortsman, Schramowski, Kundurthy, Crowson, Schmidt, Kaczmarczyk, and Jitsev]{laion}
Christoph Schuhmann, Romain Beaumont, Richard Vencu, Cade Gordon, Ross Wightman, Mehdi Cherti, Theo Coombes, Aarush Katta, Clayton Mullis, Mitchell Wortsman, Patrick Schramowski, Srivatsa Kundurthy, Katherine Crowson, Ludwig Schmidt, Robert Kaczmarczyk, and Jenia Jitsev.
\newblock Laion-5b: An open large-scale dataset for training next generation image-text models, 2022.
\newblock URL \url{https://arxiv.org/abs/2210.08402}.

\bibitem[Tschannen et~al.(2025)Tschannen, Gritsenko, Wang, Naeem, Alabdulmohsin, Parthasarathy, Evans, Beyer, Xia, Mustafa, et~al.]{siglip_2}
Michael Tschannen, Alexey Gritsenko, Xiao Wang, Muhammad~Ferjad Naeem, Ibrahim Alabdulmohsin, Nikhil Parthasarathy, Talfan Evans, Lucas Beyer, Ye~Xia, Basil Mustafa, et~al.
\newblock Siglip 2: Multilingual vision-language encoders with improved semantic understanding, localization, and dense features.
\newblock \emph{arXiv preprint arXiv:2502.14786}, 2025.

\bibitem[Tsipras et~al.(2018)Tsipras, Santurkar, Engstrom, Turner, and Madry]{robust_tradeoff}
Dimitris Tsipras, Shibani Santurkar, Logan Engstrom, Alexander Turner, and Aleksander Madry.
\newblock Robustness may be at odds with accuracy.
\newblock \emph{arXiv preprint arXiv:1805.12152}, 2018.

\bibitem[Varma et~al.(2024)Varma, Delbrouck, Chen, Chaudhari, and Langlotz]{varma2024ravl}
Maya Varma, Jean-Benoit Delbrouck, Zhihong Chen, Akshay Chaudhari, and Curtis Langlotz.
\newblock Ravl: Discovering and mitigating spurious correlations in fine-tuned vision-language models.
\newblock \emph{Advances in Neural Information Processing Systems}, 37:\penalty0 82235--82264, 2024.

\bibitem[Vasquez-Venegas et~al.(2025)Vasquez-Venegas, Wu, Sundar, Proa, Beloy, Medina, Mcnichol, Parvataneni, Kurtzman, Mirshawka, et~al.]{medical_one}
Constanza Vasquez-Venegas, Chenwei Wu, Saketh Sundar, Renata Proa, Francis~Joshua Beloy, Jillian~Reeze Medina, Megan Mcnichol, Krishnaveni Parvataneni, Nicholas Kurtzman, Felipe Mirshawka, et~al.
\newblock Detecting and mitigating the clever hans effect in medical imaging: a scoping review.
\newblock \emph{Journal of Imaging Informatics in Medicine}, 38\penalty0 (4):\penalty0 2563--2579, 2025.

\bibitem[Wah et~al.(2011)Wah, Branson, Welinder, Perona, and Belongie]{cub}
Catherine Wah, Steve Branson, Peter Welinder, Pietro Perona, and Serge Belongie.
\newblock The caltech-ucsd birds-200-2011 dataset.
\newblock 2011.

\bibitem[Wang et~al.(2024)Wang, Lin, Chen, Schmidt, Han, and Zhang]{sober_clip}
Qizhou Wang, Yong Lin, Yongqiang Chen, Ludwig Schmidt, Bo~Han, and Tong Zhang.
\newblock A sober look at the robustness of clips to spurious features.
\newblock \emph{Advances in Neural Information Processing Systems}, 37:\penalty0 122484--122523, 2024.

\bibitem[Wortsman et~al.(2022)Wortsman, Ilharco, Kim, Li, Kornblith, Roelofs, Lopes, Hajishirzi, Farhadi, Namkoong, et~al.]{wiseft}
Mitchell Wortsman, Gabriel Ilharco, Jong~Wook Kim, Mike Li, Simon Kornblith, Rebecca Roelofs, Raphael~Gontijo Lopes, Hannaneh Hajishirzi, Ali Farhadi, Hongseok Namkoong, et~al.
\newblock Robust fine-tuning of zero-shot models.
\newblock In \emph{Proceedings of the IEEE/CVF Conference on Computer Vision and Pattern Recognition}, pages 7959--7971, 2022.

\bibitem[Xu et~al.(2023)Xu, Xie, Tan, Huang, Howes, Sharma, Li, Ghosh, Zettlemoyer, and Feichtenhofer]{metaclip}
Hu~Xu, Saining Xie, Xiaoqing~Ellen Tan, Po-Yao Huang, Russell Howes, Vasu Sharma, Shang-Wen Li, Gargi Ghosh, Luke Zettlemoyer, and Christoph Feichtenhofer.
\newblock Demystifying clip data.
\newblock \emph{arXiv preprint arXiv:2309.16671}, 2023.

\bibitem[Yalavarthi et~al.()Yalavarthi, Ratha, and Govindaraju]{dial}
Bharat~Chandra Yalavarthi, Nalini~K Ratha, and Venu Govindaraju.
\newblock Label-free mitigation of spurious correlations in vlms using sparse autoencoders.
\newblock In \emph{The Fourteenth International Conference on Learning Representations}.

\bibitem[Yuksekgonul et~al.(2022)Yuksekgonul, Bianchi, Kalluri, Jurafsky, and Zou]{linear_VLM1}
Mert Yuksekgonul, Federico Bianchi, Pratyusha Kalluri, Dan Jurafsky, and James Zou.
\newblock When and why vision-language models behave like bags-of-words, and what to do about it?
\newblock \emph{arXiv preprint arXiv:2210.01936}, 2022.

\bibitem[Zaigrajew et~al.(2025)Zaigrajew, Baniecki, and Biecek]{sae}
Vladimir Zaigrajew, Hubert Baniecki, and Przemyslaw Biecek.
\newblock Interpreting clip with hierarchical sparse autoencoders, 2025.
\newblock URL \url{https://arxiv.org/abs/2502.20578}.

\bibitem[Zhang et~al.(2024)Zhang, Huang, Jin, and Lu]{clip_survey}
Jingyi Zhang, Jiaxing Huang, Sheng Jin, and Shijian Lu.
\newblock Vision-language models for vision tasks: A survey.
\newblock \emph{IEEE Transactions on Pattern Analysis and Machine Intelligence}, 46\penalty0 (8):\penalty0 5625--5644, 2024.

\bibitem[Zhang et~al.(2023)Zhang, He, Xu, Yu, Shen, and Cui]{nico++}
Xingxuan Zhang, Yue He, Renzhe Xu, Han Yu, Zheyan Shen, and Peng Cui.
\newblock Nico++: Towards better benchmarking for domain generalization.
\newblock In \emph{Proceedings of the IEEE/CVF Conference on Computer Vision and Pattern Recognition}, pages 16036--16047, 2023.

\bibitem[Zhou et~al.(2017)Zhou, Lapedriza, Khosla, Oliva, and Torralba]{places}
Bolei Zhou, Agata Lapedriza, Aditya Khosla, Aude Oliva, and Antonio Torralba.
\newblock Places: A 10 million image database for scene recognition.
\newblock \emph{IEEE Transactions on Pattern Analysis and Machine Intelligence}, 40\penalty0 (6):\penalty0 1452--1464, 2017.

\end{thebibliography}

\appendix

\section*{Appendix Overview}
\addcontentsline{toc}{section}{Appendix Overview}

Unless otherwise specified, all ablation studies utilize the CLIP ViT-B/16 architecture initialized with LAION-2B weights and are evaluated on the Waterbirds-100\% benchmark. Any deviations from this standard configuration are explicitly noted within their respective sections.

\tableofcontents

\newpage

\section{Appendix: CLIP Text Encoder Additivity}\label{app: text encoder addititivty}

\subsection{Probing Compositional Linear Superposition}

To assess whether VLM text encoders behave approximately as a bag-of-words model, we examine whether phrase embeddings are represented as a linear superposition of their constituent word embeddings. Given two content words $w_a$ (foreground) and $w_b$ (background) and an encoder $f(\cdot)$, we embed single-word prompts $s_a, s_b$ and the joint prompt $s_{a,b}$ into unit vectors $e_a, e_b, e_{a,b} \in \mathbb{R}^d$. The degree of additivity is quantified by the cosine similarity:
\[
\mathrm{Sim}_{\text{Both,Sum}}(w_a, w_b) = \cos(e_{a,b},\, e_a + e_b).
\]
High values of $\mathrm{Sim}_{\text{Both,Sum}}$ indicate that the phrase embedding is well-approximated by the vector sum of its parts, a hallmark of bag-of-words models like Word2Vec.

\subsection{Experimental Setup and Prompt Construction}

The evaluation spans 5,000 foreground--background word pairs (e.g., ``bird''--``swamp'') chosen to reflect couplings that frequently appear as spurious correlations in vision benchmarks. For VLM text encoders, prompts are instantiated using a standard image-centric template:
\begin{itemize}
    \item $s_a = \text{``a photo of a } w_a\text{''}$
    \item $s_b = \text{``a photo of a } w_b\text{''}$
    \item $s_{a,b} = \text{``a photo of a } w_a, w_b\text{''}$
\end{itemize}
For non-VLM baselines (Word2Vec and BERT), raw word strings are used (e.g., $s_a = w_a$) to avoid injecting template artifacts foreign to their pre-training domains. All resulting embeddings are $\ell_2$-normalized such that $\lVert e_a \rVert_2 = \lVert e_b \rVert_2 = \lVert e_{a,b} \rVert_2 = 1$. We report both the compositional additivity ($\mathrm{Sim}_{\text{Both,Sum}}$) and the lexical similarity between the words themselves ($\mathrm{sim}_{\text{T1,T2}}(w_a,w_b)=\cos(e_a,e_b)$).

\begin{table}[ht]
\centering
\caption{Cosine similarities (mean $\pm$ standard deviation) across 5,000 word pairs, grouped by model family.}
\label{tab:reorganized_sim}
\begin{small}
\begin{sc}
\begin{tabular}{llcc}
\toprule
\textbf{Family} & \textbf{Architecture} & $\mathrm{Sim}_{\mathrm{T1,T2}}$ & $\mathbf{Sim}_{\mathbf{B,S}}$ \\
\midrule
bag-of-words & Word2Vec &  $0.03 \pm 0.08$ & $0.99 \pm 0.01$ \\
\midrule
\multirow{3}{*}{CLIP} & ViT-B/16 & $0.61 \pm 0.06$ & $0.90 \pm 0.03$ \\
 & ViT-L/14 & $0.65 \pm 0.04$ & $0.87 \pm 0.02$ \\
 & ConvNext-W & $0.70 \pm 0.06$ & $0.91 \pm 0.02$ \\
\midrule
\multirow{2}{*}{SigLIP 2} & ViT-B/16 & $0.71 \pm 0.06$ & $0.93 \pm 0.04$ \\
 & ViT-L/16 & $0.75 \pm 0.06$ & $0.85 \pm 0.03$ \\
\midrule
Contextual& BERT  & $0.66 \pm 0.12$ & $0.79 \pm 0.03$ \\
\bottomrule
\end{tabular}
\end{sc}
\end{small}
\end{table}

Results in Table \ref{tab:reorganized_sim} mirror the vision encoders' in Table \ref{tab:image_additivity}. As a baseline, Word2Vec is almost perfectly additive ($\mathrm{Sim}_{\text{Both,Sum}} \approx 1.0$), while BERT is substantially less so ($0.79 \pm 0.03$). This discrepancy reflects BERT's deep contextualization and self-attention mechanism, where the embedding of a phrase like ``duck, pond'' is a non-linear, contextually enriched representation rather than a simple superposition of constituents. 

In contrast, VLM text encoders remain highly additive, nearly matching the bag-of-words behavior of Word2Vec. This suggests that VLM contrastive training objective encourages the detection of salient keywords over the modeling of complex phrasal structure. 

This near-linear superposition has direct implications for background bias. When $e_{a,b}$ is well-approximated by $e_a + e_b$, foreground and background components occupy nearly independent axes in the text embedding space. Consequently, the background direction remains prominent and can be reused across disparate objects (e.g., ``bird in swamp'' vs. ``dog in swamp''). In the multimodal alignment space, images sharing a background but differing in object will likely end up close to one another whenever the background component is strong, facilitating CLIP's reliance on backgrounds as classification shortcuts. These results mirror our findings on the image-encoder side, Section \ref{sec:image encoder additivity}, suggesting a consistent mechanistic behavior where VLMs represent scenes as a "bag of features" across both modalities.

\section{Additional Results}\label{app:extra_results}

We show here some additional evaluations that did not have enough room in the main body. The first evalutation is on the Waterbirds benchmark using a CLIP model with a ConvNeXT-W backbone with results shown in Table \ref{tab:waterbirds_convnext}. Performance gains made by BAP are very similar to those observed in Table \ref{tab:waterbirds} indicating cross-architectural stability.

\begin{table}[ht]
  \caption{Average and worst-group accuracies (mean $\pm$ std over 5 runs) across various baselines for the Waterbirds benchmark using 95\% and 100\% spurious correlation rates. BAP outperforms all other baselines and retains near-identical robustness under both 95\% and 100\% domains.}
  \label{tab:waterbirds_convnext}
  \centering
  \setlength{\tabcolsep}{6pt} 
  \begin{small}
  \begin{tabular}{l cccc}
  \toprule
  & \multicolumn{4}{c}{\textbf{CLIP | ConvNeXT-W}} \\
  \cmidrule(lr){2-5}
  & \multicolumn{2}{c}{\textbf{Waterbirds-95\%}} & \multicolumn{2}{c}{\textbf{Waterbirds-100\%}} \\
  \cmidrule(lr){2-3} \cmidrule(lr){4-5}
  \textbf{Method} & \textbf{AVG} ($\uparrow$) & \textbf{WGA} ($\uparrow$) & \textbf{AVG} ($\uparrow$) & \textbf{WGA} ($\uparrow$) \\
  \midrule
  \multicolumn{5}{c}{\textit{\textbf{Zero-Shot Evaluation (No Downstream Labels)}}} \\
  \midrule
  Native Backbone & $75.3 \std{0.0}$ & $47.2 \std{0.0}$ & $75.3 \std{0.0}$ & $47.2 \std{0.0}$ \\
  RoboShot & $75.8 \std{0.0}$ & $59.2 \std{0.0}$ & $75.8 \std{0.0}$ & $59.2 \std{0.0}$ \\
  DIAL & $79.7 \std{0.0}$ & $64.9 \std{0.0}$ & $79.7 \std{0.0}$ & $64.9 \std{0.0}$ \\
  \textbf{BAP (ZS)} (Ours) & $\textbf{86.4} \std{0.0}$ & $\textbf{71.6} \std{0.0}$ & $\textbf{86.4} \std{0.0}$ & $\textbf{71.6} \std{0.0}$ \\
  \midrule
  \multicolumn{5}{c}{\textit{\textbf{Trained Evaluation (Linear Probing \& Fine-Tuning)}}} \\
  \midrule
  Native Backbone (LP) & $81.5 \std{0.7}$ & $57.1 \std{1.1}$ & $63.3 \std{0.4}$ & $31.3 \std{1.0}$ \\
  AFR & $92.9 \std{0.3}$ & $88.9 \std{1.9}$ & - & - \\
  DFR & $\textbf{94.1} \std{0.4}$ & $91.3 \std{0.5}$ & - & - \\
  ERM (LP-FT) & $87.5 \std{0.5}$ & $73.9 \std{0.6}$ & $63.1 \std{0.8}$ & $18.7 \std{1.4}$ \\
  WiSE-FT & $88.4 \std{0.4}$ & $65.8 \std{0.5}$ & $63.1 \std{0.2}$ & $21.6 \std{0.6}$ \\
  PruSC & $75.7 \std{3.0}$ & $57.5 \std{3.7}$ & $47.3 \std{2.9}$ & $35.6 \std{4.5}$ \\
  Data-Matched Control & $92.5 \std{1.2}$ & $85.8 \std{3.8}$ & $87.2 \std{2.5}$ & $80.7 \std{5.2}$ \\
  \textbf{BAP (LP)} (Ours) & $93.6 \std{0.8}$ & $\mathbf{92.5 \std{1.0}}$ & $\mathbf{94.0 \std{0.6}}$ & $\mathbf{91.9 \std{1.2}}$ \\
  \bottomrule
  \end{tabular}
  \end{small}
\end{table}

The second set of results complement the vehicle classification task in Table \ref{tab:nico++ results} and were obtained using a SigLIP 2 model with a Vit-B/16 unlike the CLIP ConvNeXT shown in the main body. Results are visible in Table \ref{tab:nico_vit} and indicate that performance using SigLIP 2 is consistent with CLIP results seen in the main body.  This further set of evaluations strongly supports our method's ability to generalize to real-world data.

\begin{table}[ht]
\centering
\caption{Average and worst-group accuracies (mean $\pm$ std over 5 runs) across the best performing baselines for four pairs of binary vehicle classification on the NICO++ dataset using SigLIP 2 ViT-B/16. }
\label{tab:nico_vit}
\small
\setlength{\tabcolsep}{2.3pt}
\begin{tabular}{l cccccccc}
\toprule
& \multicolumn{2}{c}{\textbf{Car vs. Truck}}  
& \multicolumn{2}{c}{\textbf{Car vs. Bus}}  
& \multicolumn{2}{c}{\textbf{Ship vs. Sailboat}}  
& \multicolumn{2}{c}{\textbf{Bike vs. Motorbike}} \\
\cmidrule(lr){2-3} \cmidrule(lr){4-5} \cmidrule(lr){6-7} \cmidrule(lr){8-9}
\textbf{Method} 
& \textbf{AVG} ($\uparrow$) & \textbf{WGA} ($\uparrow$) 
& \textbf{AVG} ($\uparrow$) & \textbf{WGA} ($\uparrow$) 
& \textbf{AVG} ($\uparrow$) & \textbf{WGA} ($\uparrow$)
& \textbf{AVG} ($\uparrow$) & \textbf{WGA} ($\uparrow$) \\
\midrule
Native Backbone (ZS) 
& 85.2 \std{0.0} & 59.4 \std{0.0} 
& \textbf{88.9} \std{0.0} & 61.7 \std{0.0} 
& 77.6 \std{0.0} & 55.2 \std{0.0}
& 81.4 \std{0.0} & 49.2 \std{0.0} \\

DIAL 
& 83.5 \std{0.0} & 64.1 \std{0.0} 
& 87.1 \std{0.0} & 64.9 \std{0.0} 
& \textbf{79.3} \std{0.0} & \textbf{59.8} \std{0.0}
& 77.0 \std{0.0} & 52.7 \std{0.0} \\

\textbf{BAP (ZS)} (Ours) 
& \textbf{86.1} \std{0.0} & \textbf{69.4} \std{0.0} 
& 87.4 \std{0.0} & \textbf{71.3} \std{0.0} 
& 77.2 \std{0.0} & 58.1 \std{0.0}
& \textbf{77.9} \std{0.0} & \textbf{60.4} \std{0.0} \\

\midrule
\midrule

Native Backbone (LP) 
& 80.5 \std{1.2} & 31.2 \std{3.4} 
& 85.2 \std{0.5} & 38.6 \std{1.1} 
& 75.5 \std{0.4} & 29.9 \std{1.2}
& 78.3 \std{1.1} & 24.9 \std{1.8} \\

Data-Matched Control 
& 83.4 \std{0.8} & 57.1 \std{2.6} 
& 86.9 \std{0.4} & 61.8 \std{0.7} 
& \textbf{84.1} \std{0.6} & 59.9 \std{1.5}
& 88.6 \std{1.5} & 64.3 \std{2.0} \\

\textbf{BAP (LP)} (Ours) 
& \textbf{84.7} \std{1.7} & \textbf{65.2} \std{2.3} 
& \textbf{88.2} \std{1.8} & \textbf{70.3} \std{2.5} 
& 83.0 \std{1.4} & \textbf{72.9} \std{2.9}
& \textbf{90.2} \std{1.9} & \textbf{74.1} \std{3.2} \\

\bottomrule
\end{tabular}
\end{table}

\section{Additional backbone testing}\label{app:backbone_ablation}
\subsection{Ablation on Backbone Size}

To determine the sensitivity of our approach to model capacity, we performed an ablation study evaluating different backbone sizes (ViT-B/16, ViT-L/14, and ViT-H/14) for both CLIP and SigLIP 2 on the Waterbirds-100\% task. The results, detailing the mean and standard deviation over multiple runs, are presented in Table~\ref{tab:backbone_ablation}. 

The consistent performance of BAP across various backbone sizes for both CLIP and SigLIP 2 indicates that it is a highly robust algorithm. While scaling the backbone from a Base to a Huge architecture yields slight improvements in overall accuracy, BAP successfully maintains peak worst-group accuracy exceeding $90\%$ across the board. This highlights that our method effectively reshapes the inductive bias of the representation space independently of the underlying parameter count or architectural scale.

\begin{table}[h]
\centering
\caption{Average and worst-group accuracies (mean $\pm$ std over 3 runs) for CLIP and SigLIP 2 across different backbone sizes on the Waterbirds-100\% benchmark.}
\label{tab:backbone_ablation}
\begin{tabular}{lcccc}
\toprule
& \multicolumn{2}{c}{\textbf{CLIP}} & \multicolumn{2}{c}{\textbf{SigLIP 2}} \\
\cmidrule(lr){2-3} \cmidrule(lr){4-5}
\textbf{Backbone} & \textbf{AVG ($\uparrow$)} & \textbf{WGA ($\uparrow$)} & \textbf{AVG ($\uparrow$)} & \textbf{WGA ($\uparrow$)} \\
\midrule
ViT-B/16 & 93.40 \std{0.59} & 90.58 \std{0.77} & 95.67 \std{0.25} & 92.37 \std{0.32} \\
ViT-L/14 & 95.96 \std{0.10} & 93.69 \std{0.51} & 97.77 \std{0.12} & 95.25 \std{0.21} \\
ViT-H/14 & 96.79 \std{0.30} & 93.99 \std{0.43} & 98.18 \std{0.18} & 95.57 \std{0.59} \\
\bottomrule
\end{tabular}
\end{table}

\subsection{Ablation on Pre-trained Checkpoints}

To verify that the robustness gains achieved by BAP are not reliant on a specific pre-training data distribution, we conducted an ablation study across several distinct pre-trained checkpoints for the CLIP ViT-B/16 architecture. We evaluated the model using weights derived from LAION, OpenAI, MetaCLIP, and DataComp \citep{laion, CLIP_original, metaclip, datacomp}.

The results, detailed in Table~\ref{tab:checkpoint_ablation}, demonstrate the stability of BAP. Notably, the Data-Matched ERM Control yields highly variable performance, with Worst-Group Accuracy (WGA) fluctuating drastically between $63.64\%$ and $87.38\%$ depending on the pre-training dataset. In contrast, BAP remains remarkably consistent, maintaining a WGA above $91\%$ regardless of the underlying pre-trained checkpoint. This consistent performance across various base weights indicates that BAP is a highly robust algorithm whose background invariance mechanisms generalize effectively regardless of the initial representation space.

\begin{table}[h]
\centering
\caption{Average and worst-group accuracies (mean $\pm$ std) for various CLIP ViT-B/16 pre-trained checkpoints. The Native Representations (LP) baseline is included for comparison. BAP demonstrates consistent robustness across all weights, whereas the Data-Matched Control exhibits high variance.}
\label{tab:checkpoint_ablation}
\setlength{\tabcolsep}{5pt} 
\small
\begin{tabular}{lcccccc}
\toprule
& \multicolumn{2}{c}{\textbf{Native Reps. (LP)}} & \multicolumn{2}{c}{\textbf{Data-Matched Control}} & \multicolumn{2}{c}{\textbf{BAP (Ours)}} \\
\cmidrule(lr){2-3} \cmidrule(lr){4-5} \cmidrule(lr){6-7}
\textbf{Pre-training Checkpoint} & \textbf{AVG ($\uparrow$)} & \textbf{WGA ($\uparrow$)} & \textbf{AVG ($\uparrow$)} & \textbf{WGA ($\uparrow$)} & \textbf{AVG ($\uparrow$)} & \textbf{WGA ($\uparrow$)} \\
\midrule
LAION (s34b\_b88k) & 60.46 \std{1.2} & 19.56 \std{3.3} & 83.34 \std{2.3} & 64.12 \std{1.7} & \textbf{93.67 \std{0.7}} & \textbf{92.21 \std{4.3}} \\
OpenAI & 62.75 \std{2.1} & 20.62 \std{3.7} & 92.87 \std{1.1} & 87.38 \std{1.9} & \textbf{93.84 \std{3.3}} & \textbf{91.12 \std{4.8}} \\
MetaCLIP (fullcc) & 62.75 \std{1.7} & 22.26 \std{2.0} & 91.91 \std{1.4} & 83.86 \std{1.7} & \textbf{94.18 \std{1.7}} & \textbf{92.06 \std{2.6}} \\
DataComp-XL (s13b\_b90k) & 61.56 \std{2.5} & 20.09 \std{3.1} & 83.53 \std{1.8} & 63.64 \std{2.2} & \textbf{93.20 \std{0.9}} & \textbf{91.84 \std{1.2}} \\
\bottomrule
\end{tabular}
\end{table}

\section{Embedding Space Analysis}\label{app: BSI and UMAP}
\subsection{Background Sensitivity Index (BSI)}

In order to assess the effect of our method on VLM embedding spaces, we introduce a proprietary metric aimed at quantifying the degree of background invariance. The Background Sensitivity Index (BSI) quantifies a vision encoders's sensitivity to background changes by measuring the shift in object representations relative to their natural embedding variance. For a fixed object class, we partition the vision encoder' embeddings \(\mathbf{z}\) by background into sets \(A\) and \(B\), compute centroids \(\boldsymbol{\mu}_A, \boldsymbol{\mu}_B\) and variances \(\sigma_A^2, \sigma_B^2\), then calculate
\[
\text{BSI} = \frac{\|\boldsymbol{\mu}_A - \boldsymbol{\mu}_B\|_2}{\sqrt{\sigma_A^2 + \sigma_B^2}}.
\]

Specifically, for the Waterbirds-100\% benchmark, we calculate this index by isolating the impact of the background on each species' representation. For the landbird class, we define set $A$ using using landbird images placed against land backgrounds and set $B$ using the same bird images with the land backgrounds swapped for water ones. We then compute the respective centroids and variances ($\boldsymbol{\mu}_A, \sigma_A^2$ and $\boldsymbol{\mu}_B, \sigma_B^2$) for these sets and apply the BSI equation. This process is repeated for the waterbird class, and the results are averaged to provide a unified measure of how much the background swap, regardless of the bird itself, shifts the model's internal representations.  Results for both CLIP and SigLIP 2 are presented in Table \ref{tab: BSI}.

\begin{table*}[ht]
\centering
\caption{Background Sensitivity Index (BSI) (mean $\pm$ std) for Waterbirds-100\%  across  a variety of baselines.}
\label{tab: BSI}
\begin{small}
\begin{tabular}{ll|c|c}
\toprule
& & \multicolumn{1}{c}{\textbf{CLIP}} & \multicolumn{1}{c}{\textbf{SigLIP 2}} \\
\cmidrule(lr){3-3} \cmidrule(lr){4-4}
\textbf{Method} & & \textbf{BSI}($\downarrow$) & \textbf{BSI} ($\downarrow$)\\
\midrule
\midrule
Native Backbone (LP) & & $30.4 \std{0.0}$ & $36.8 \std{1.3}$ \\
ERM (LP-FT) & & $38.5 \std{1.2}$ & $45.1 \std{3.5}$ \\
PruSC & & $31.4 \std{2.1}$ & $34.9 \std{1.7}$ \\
Data-Matched Control & & $9.8 \std{0.8}$ & $12.3 \std{0.7}$ \\
\textbf{BAP (LP)} (Ours) & & $\mathbf{2.2} \std{0.1}$ & $\mathbf{2.5} \std{0.3}$ \\
\midrule
\bottomrule
\end{tabular}
\end{small}
\end{table*}

Low BSI values indicate strong background invariance, where centroid shifts remain small compared to intra-group spread; high values reveal notable sensitivity, where background changes drive substantial embedding shifts. As can be seen in Table \ref{tab: BSI}, BAP  facilitates a level of background invariance that is unmatched by any of the other baselines.  Our approach is an order of magnitude lower in background sensitivity compared to VLM native representations and nearly three times lower than the Data-Matched control, further indicating that our anchor generation and alignment strategy facilitates background invariance in a manner not possible using standard cross-entropy. 

\begin{figure*}[ht]
    \centering
    \hspace*{0.035\textwidth} 
    \begin{minipage}{0.96\textwidth}
        \centering
        \includegraphics[trim={5cm} {4.5cm} {4.5cm} {0.5cm}, clip, width=\linewidth]{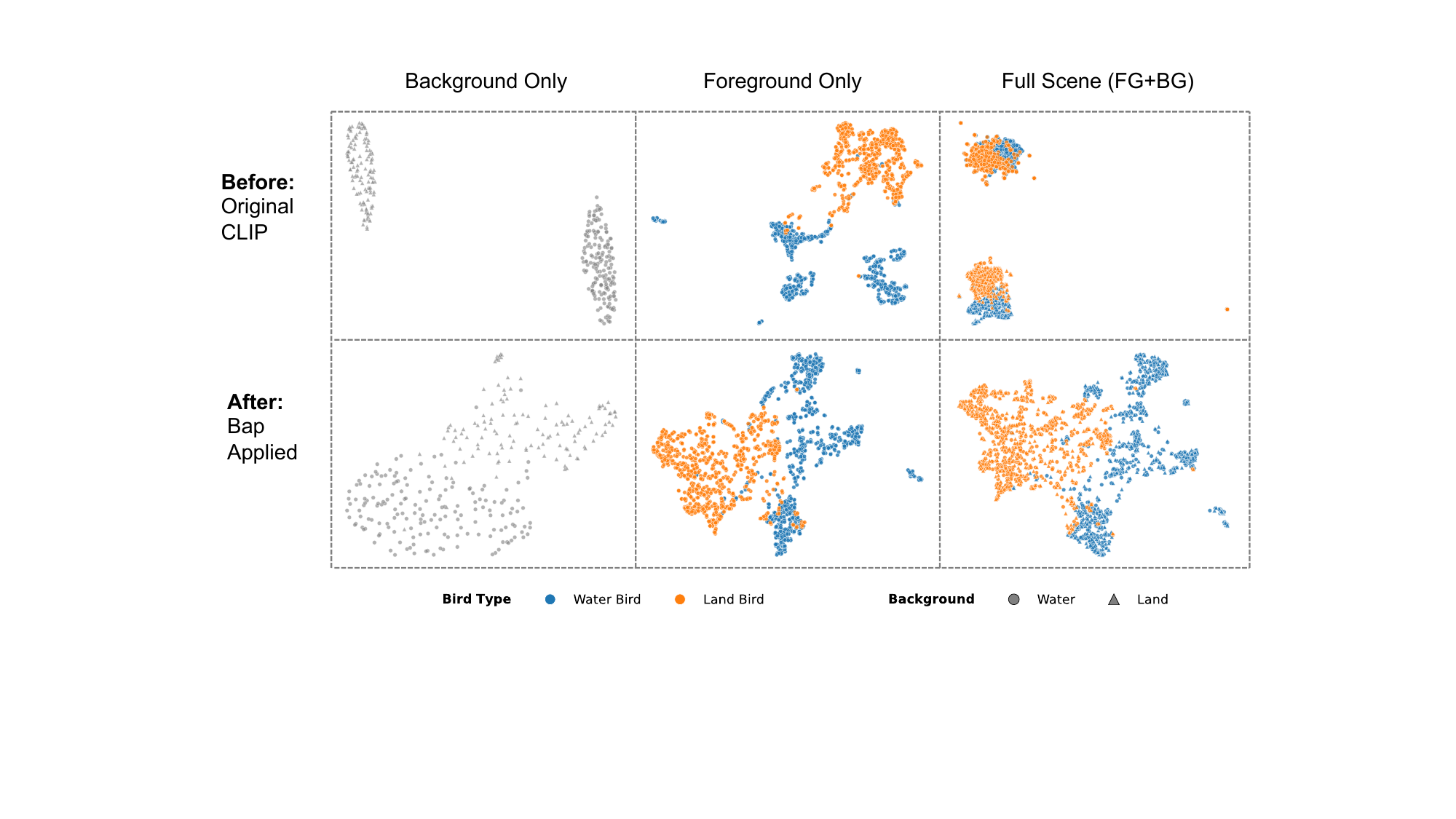}
            \caption{\textbf{UMAP visualizations of a vision encoders embedding space before and after BAP.}  The top row contains UMAP visualizations for CLIP's  embedding space while the bottom row contains visualizations after BAP application. From left to right we have: background-only  embeddings, foreground-only embeddings, and full scene ( foregrounds composited onto backgrounds ) embeddings. }
            \label{fig:umap}
\end{minipage}
\end{figure*}

\subsection{UMAP visualization of embedding space }
To visualize the effect that BAP has on the embedding space, we plot UMAP visualizations for isolated foregrounds, backgrounds, and composite scenes using CLIP embeddings; we show both native CLIP embeddings and post-BAP embeddings in Figure \ref{fig:umap}. Most notably, in the full scene embeddings, the original CLIP representation shows a hierarchical grouping where images are first grouped by background and then sub-grouped by bird class. However, post-BAP embeddings indicate that the separation is made almost entirely based on the foreground class. 

Note that the separation in BAPs background-only embeddings is significantly decreased from the original CLIP embeddings.  Our investigation reveals that  BAP negatively impacts background and scene recognition, for further details, see Appendix \ref{app: bap zero shot}.

\section{Baseline selection justification}\label{app: baselines}

In addition to BAP and the data-matched control, we compare our method against a variety of general and VLM-specific robustness interventions to comprehensively assess performance:

\begin{itemize}
    \item \textbf{Native Representations:} We evaluate the original pre-trained vision encoders in both zero-shot and linear probing capacities. Both provide a starting point to assess the native model's baseline representations.
    
    \item \textbf{ERM (LP-FT):} We perform Empirical Risk Minimization using a standard linear probing then fine-tuning approach \citep{lp_then_ft}.
    
    \item \textbf{Last-Layer Retraining:} We utilize Deep Feature Reweighting (DFR) \citep{dfr} as a representative oracle baseline, alongside Automatic Feature Reweighting (AFR) \citep{afr} as a representative non-oracle baseline.
    
    \item \textbf{Representation-Level Interventions:} We employ PruSC \citep{prusc} as a general state-of-the-art representation-level intervention.
    
    \item \textbf{VLM-Specific Adaptations:} We compare against WiseFT \cite{wiseft}, which achieves robustness through the weight-space ensembling of zero-shot and fine-tuned models, as well as RoboShot \citep{roboshot}, which uses spurious feature insights to adapt model representations through embedding space operations.
    
    \item \textbf{Sparse Autoencoders:} Lastly, we evaluate against DIAL \citep{dial}, a state-of-the-art label-free baseline that removes spurious correlations from dense VLM embeddings using sparse autoencoders.
\end{itemize}

\section{Appendix: Ablation on Anchor Semantic Necessity}\label{app: orthogonal targets}

\paragraph{Random Orthogonal Target Assignment}
In the standard BAP formulation, Phase 1 is designed to extract a set of foreground-only anchor vectors using a frozen teacher encoder. This process aims to preserve the rich semantic structure of foreground objects while suppressing background noise. To isolate whether the observed robustness gains are driven by these high-fidelity semantic anchors or simply by the geometric enforcement of invariance, we conducted an ablation study using random orthogonal training targets.

This ablation entirely removes \textbf{Phase 1: Anchor Vector Extraction}. Instead of computing unique, instance-specific anchor vectors ($\mathbf{a}$) by averaging composites, we initialize two static, synthetic target vectors: $\mathbf{v}_{water}$ and $\mathbf{v}_{land}$.

\begin{itemize}
    \item Initialization: The vectors $\mathbf{v}_{water}$ and $\mathbf{v}_{land}$ are generated as random orthogonal vectors in the embedding space $\mathbb{R}^d$ and normalized to the unit hypersphere such that $\|\mathbf{v}\|_2 = 1$ and $\mathbf{v}_{water} \cdot \mathbf{v}_{land} = 0$.
    \item Target Assignment: We collapse the granular species-level distinctiveness of the training targets. All foreground instances belonging to the Waterbird super-class are assigned $\mathbf{v}_{water}$ as their target, while all Landbird instances are assigned $\mathbf{v}_{land}$.
    \item Alignment Training: We then proceed to \textbf{Phase 2: Robust Alignment Pre-training}. The student encoder $f_{\theta}$ is optimized to minimize the cosine distance between the embedding of a randomized synthetic composite and its assigned super-class vector:
    \begin{equation}
\mathcal{L}_{align}
=
1
-
\frac{
f_{\theta}(\hat{x})^{\top}\mathbf{v}_{target}
}{
\left\|f_{\theta}(\hat{x})\right\|_2
\left\|\mathbf{v}_{target}\right\|_2
}
\end{equation}
\end{itemize}

This configuration maintains the same "virtual epoch" size and data exposure as the standard BAP protocol. However, it explicitly discards the learned semantic foreground signal of specific species (e.g., "Black-footed Albatross") in favor of a coarse super-class grouping. By comparing this baseline against standard BAP, we can quantify the degree to which robustness relies on the high-fidelity distillation of the foreground signal versus the simple, forced suppression of background features.  For this ablation, we used the exact same CUB-Places365 paradigm as in Sections \ref{sec: results} to conduct pre-training.  Following this, we train various linear probes on a number of downstream data sets, again identical to deployment in the main body; we tested downstream performance on the Waterbirds benchmark followed by two CounterAnimal pairs.  Note that we consider the Waterbirds task to be an in-distribution (I.D) task, since the same CUB-Places365 paradigm is used in both training and testing,  while we consider the CounterAnimal tasks as O.O.D tasks.  

\begin{table}[h]
    \centering
    \caption{Performance Metrics for Random Orthogonal Target Alignment. We report Average Accuracy and Worst Group Accuracy across different evaluation tasks (mean $\pm$ std over 3 runs).}
    \label{tab:orthogonal_results}
    \begin{tabular}{lcc}
        \toprule
        \textbf{Task} & \textbf{AVG} & \textbf{WGA} \\
        \midrule
        \textbf{Waterbirds} & 93.5 \std{0.5} & 90.2 \std{0.4} \\
        \textbf{CounterAnimal}: Bulbul vs. Brambling & 23.1 \std{4.1} & 11.7 \std{3.7} \\
        \textbf{CounterAnimal}: Ptarmigan vs. Prairie-Chicken & 14.7 \std{5.3} & 7.1 \std{2.7} \\
        \bottomrule
    \end{tabular}
\end{table}

The results in Table \ref{tab:orthogonal_results} tell a very compelling story: for the I.D task (Waterbirds) we see nearly identical performance to the full BAP implementation in Table \ref{tab:waterbirds}, however, performance collapses catastrophically for the O.O.D CounterAnimal tasks.

The success of this ablation on the I.D Waterbirds task indicates that the mere projection of all Waterbird and Landbird instances to single fixed vectors is sufficient to suppress the various background signals. Indeed, we believe that this behavior can best be understood through the lens of an information bottleneck and by analogy to a $\beta$-VAE \citep{beta-VAE}. In a $\beta$-VAE, a weighted KL-divergence term creates a tight information bottleneck that forces the model to encode only the factors of variation necessary to reconstruct the shared structure of the data, effectively pruning out non-essential details. Similarly, the $\mathcal{L}_{align}$ objective enforces a strict many-to-one mapping. By constraining the model to map several distinct background variations of the same object to a singular, invariant point a in the embedding space, we artificially induce a bottleneck. To minimize the loss, the encoder is forced to discard the high-variance, spurious background information and encode only the salient, common factors pertaining to the birds in this instance.

However, catastrophic performance in O.O.D tasks makes this ablated version of BAP  unusable in the real world. We believe that  the drop in performance is due to two factors:  firstly, the above formulation's information bottleneck is restrictive to the point of potentially causing catastrophic forgetting  where the vision encoder is incapable of recognizing O.O.D samples. Secondly, our formulation of the anchor vectors preserves the VLM's rich semantic instance-specific information, while this version discards it leading to a potentially irregular embedding space that is void of the rich semantic structure that VLMs are known for.

The above ablation isolates the necessity of different aspects of BAP:

\begin{enumerate}
    \item Mapping several instances of a given foreground, on randomized backgrounds, to a single point in the embedding space is responsible for the suppression of background signals and produces the observed quality of background invariance.
    \item  Utilizing a frozen VLM vision encoder as a teacher model to create  individualized anchor vectors preserves semantic information  which allows the increased robustness to transfer to O.O.D domains. Furthermore, this aspect facilitates a more gentle restructuring of the embedding space, and by extension the internal model representations, such that catastrophic forgetting is mitigated.   
\end{enumerate}

\section{Experimental Details for the $K$-Ablation Analysis}
\label{app:k_ablation_appendix}

This section details the specific experimental protocol used to generate the results evaluating the isolation of the foreground signal and the attenuation of background noise as the number of composites, $K$, increases seen in Figure (1). The script performs a dual evaluation across both CLIP and SigLIP 2 architectures.

\subsection{Model Initialization and Text Embedding}
The experiment evaluates two vision-language models: OpenCLIP's ViT-B-16 (pretrained on LAION-2B) and SigLIP 2 (base-patch16-224). For both models, robust text embeddings were pre-computed for the target foreground classes and the 30 background categories. These text embeddings were generated by ensembling the tokens across 40 diverse prompt templates (e.g., "a photo of a \{\}.", "a blurry photo of a \{\}.") and normalizing the resulting mean vector.

\subsection{Estimation of the True Population Mean ($\mathbf{\mu_{bg}}$)}
To empirically calculate the residual variance, $Var(\mathbf{\epsilon})$, we first established a baseline estimation of the true background population mean, $\mathbf{\mu_{bg}}$. This was achieved by passing the 100,00 of pure Places365 background images through both the CLIP and SigLIP 2 vision encoders in large batches. The resulting normalized embeddings were averaged to compute $\mathbf{\mu_{bg}}$ for each model.

\subsection{Composite Generation and Optimization}
For each valid COCO instance, the foreground object was cropped using its bounding box and isolated using its corresponding segmentation mask. To optimize the compositing process, the isolated foreground and its mask were scaled once to $80\%$ of the target $224 \times 224$ resolution. 

The script then iteratively pasted this single optimized foreground onto up to $K_{max} = 40$ randomly sampled background images from the RAM cache. The specific progression of $K$ values tracked during the experiment were $K \in \{1, 2, 3, 5, 8, 10, 15, 20, 30, 40\}$.

\subsection{Inference and Metric Calculation}
The generated composites, alongside the pure background images used to create them, were converted to normalized tensors and passed through the vision encoders. For each step $K$:
\begin{itemize}
    \item \textbf{Purified Vector:} The embeddings of the first $K$ composites were averaged and $L_{2}$-normalized to create the purified vector.
    \item \textbf{Cosine Similarities:} The purified vector's cosine similarity to the ensembled foreground text embedding was calculated. Simultaneously, the maximum cosine similarity between the purified vector and all 30 background text embeddings was recorded].
    \item \textbf{Residual Variance:} The sample mean of the first $K$ pure background embeddings was calculated. The residual variance was then tracked by computing the squared $L_{2}$ distance between this sample mean and the pre-computed true population mean, $\mu_{bg}$
\end{itemize}

\section{Data Requirements}\label{app: data requirements}

This Section aims to assess how data-hungry BAP is  and what  the overall data requirements are in order to obtain optimal performance.
\subsubsection{$N$ hyperparameter ablation}\label{app: segmented foreground number needed}

Firstly, we study the relationship between the number of distinct foreground items $N$, used during the second phase of BAP (alignment), and performance.  Typically, we would set a high value of $M$ such that each foreground item would be repeatedly composited onto a large number of randomly selected backgrounds. However, in order to independently assess the effect of the total number of foreground items used, we opt to only repeat each foreground item twice for the purposes of this investigation ($M =2$). 

\begin{figure*}[h]
    \centering
    \hspace*{0.035\textwidth} 
    \begin{minipage}{0.96\textwidth}
        \centering
        \includegraphics[width=\linewidth]{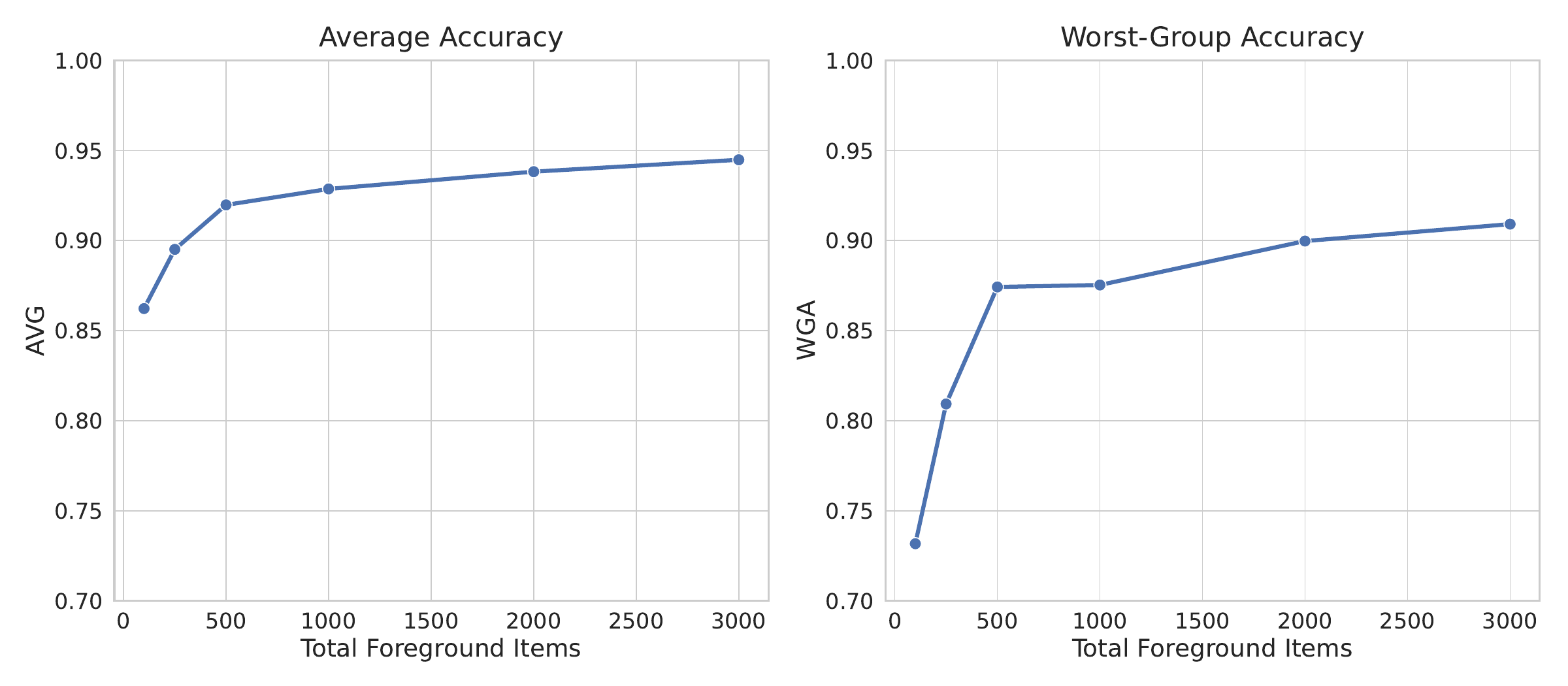}
            \caption{\textbf{Performance vs $N$ Total Segmented Foregrounds Used.} Above, we investigate the effect of the number of selected foregrounds (birds) used by BAP on performance on the Waterbirds benchmark.  The x-axis indicates how many total birds were sampled for each BAP run and the y-axes indicates average and worst group accuracy performance. We observe WGA increases sharply when going from $100$ to $500$ total birds  but then plateaus and increases slowly before reaching peak performance at around $3,000$ total birds.  }
            \label{fig:app_foreground_count}
\end{minipage}
\end{figure*}

Figure \ref{fig:app_foreground_count}, shows a rapid initial increase in worst-group accuracy, followed by a much more gradual increase to peak performance. Quite surprisingly,  we find that only $N = 500$ segmented birds was sufficient to reach $90\%$ of peak worst-group accuracy. Furthermore, we note the performance levels at  around $N = 2,000$ total foreground items nearly matches the performance reached in Table \ref{tab:waterbirds}.

\subsubsection{$K$ hyperparameter effect }\label{app: phase 1 background no.  requirements }

To assess the effect of the $K$ hyperparameter (indicating the number of composites used in the anchor generation phase) directly on model performance (rather than via similarity with corresponding text embedding as in Section \ref{sec: algo}), we construct a hyperparameter sweep. We vary the number $k$ of novel background contexts: each foreground item appears against during the anchor vector construction phase and proceeds with BAP using CUB-Places 365 composites and  the Waterbirds benchmark for linear probe training and testing. Essentially, we are varying the $K$ parameter in Equation (\ref{eq:anchor avg}) to assess its effect on training performance. 

\begin{table}[ht]
    \centering
    \caption{\textbf{Sensitivity Analysis of K.} We report Average Accuracy and Worst Group Accuracy across different values of K to determine the number of background contexts required in anchor vector generation for optimal performance .}
    \label{tab:phase 1 k vs accuracy}
    \begin{tabular}{ccc}
        \toprule
        $K$\textbf{: No. of Backgrounds} & \textbf{Average Accuracy} & \textbf{Worst Group Accuracy} \\
        \midrule
        1  & $93.5 \std{0.8}$ & $87.1 \std{1.2}$ \\
        2  & $93.9 \std{0.7}$ & $89.2 \std{0.8} $ \\
        4  & $94.1 \std{0.6}$ & $90.9 \std{0.7}$ \\
        8  & $94.7 \std{0.5}$ & $91.8 \std{0.8}$ \\
        16 & $94.5 \std{0.5}$ & $92.1 \std{0.9}$ \\
        \bottomrule
    \end{tabular}
\end{table}

We observe in Table \ref{tab:phase 1 k vs accuracy} a predictable pattern of increased random foreground-background pairings $K$  resulting in higher robustness and more stable performance. These results are in line with observations from Figure \ref{fig:k_ablation} where we demonstrated that higher $k$ values resulted in higher cosine similarity scores with the corresponding foreground specific text prompt.

\section{BAP's  practicality: full methodology and SigLIP2 results}\label{app:practical}
In Section \ref{sec:conclusion} we discuss the practical aspects of BAP and demonstrate that BAP for CLIP is highly practical because it requires only a small number of foreground objects and is robust to imperfect segmentation quality.  This section takes a deeper dive into the experimental setup for the CLIP results in Table \ref{tab:seg_conc} and Figure \ref{fig:m_conc} and presents results for SigLIP2 using the same experimental setup.

\subsection{M-hyperparameter ablation }

\begin{figure}[ht]
    \centering
    \includegraphics[width=0.8\textwidth]{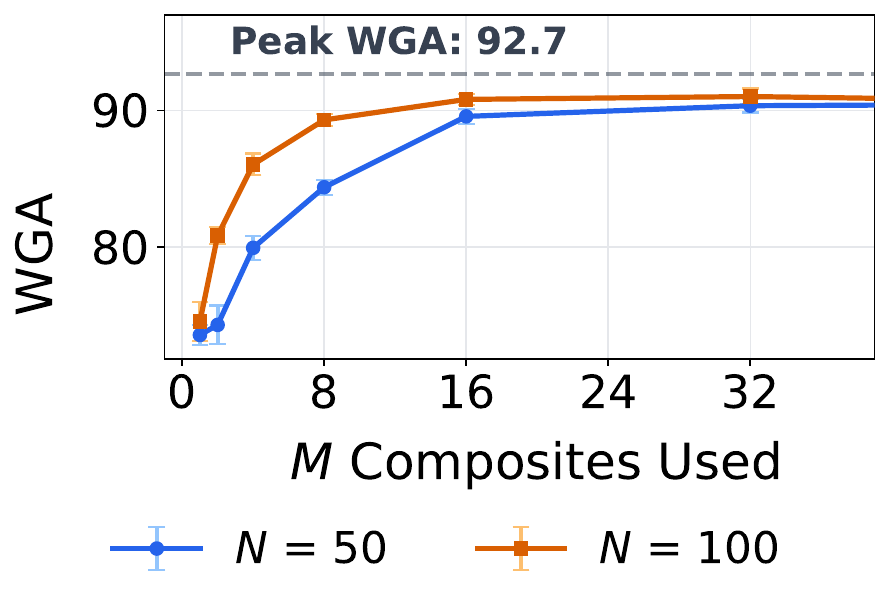}
    \caption{Impact of background randomization scaling  on model robustness under data-scarce conditions. The plot illustrates the progression of Worst-Group Accuracy (WGA) as the number of random background contexts ($M$) increases, evaluated at limited unique foreground instances ($N=50$ and $N=100$). Consistent with CLIP observation, a sharp initial improvement in WGA is observed at low $M$ values, which subsequently plateaus near peak performance for $M \ge 16$. This diminishing-returns relationship demonstrates that extensive background randomization can effectively compensate for a strict deficit in diverse foreground data.}
    \label{fig:siglip2_M_sweep}
\end{figure}

In the evaluation of the Background-invariant Anchor Pre-training (BAP) method, the $M$ hyperparameter dictates the number of distinct, randomly sampled background contexts onto which each foreground instance is composited during the robust alignment stage (Phase 2). To investigate the data-efficiency of BAP and simulate data-scarce deployment scenarios, an ablation was conducted on this parameter under restricted foreground availability (specifically $N=50$ and $N=100$ distinct items). The primary objective was to determine whether a deficit in unique, high-quality foreground segmentations could be effectively compensated for by artificially expanding the diversity of synthetic background contexts during alignment. We presented CLIP results in Figure \ref{fig:m_conc} and show SigLIP 2 results in Figure \ref{fig:siglip2_M_sweep}.

The ablation results for SigLIP 2 demonstrate that the pattern observed for CLIP in Figure \ref{fig:m_conc} is consistent across both model types which is strong evidence for BAP's high practicality and frugality as regards data requirements. 

\subsection{Ablation Study: Sensitivity to Segmentation Quality}
\label{app:segmentation_ablation}

To assess the robustness of our method against imperfect data preprocessing, we conducted a sensitivity analysis on the quality of the segmentation masks used during the synthetic composite generation. While our primary experiments utilize the high-quality, human-annotated segmentation masks provided by the CUB dataset, real-world deployment scenarios often rely on automated segmentation models which may produce noisy or coarse outputs. We therefore replicated BAP across four distinct levels of segmentation fidelity to measure the downstream impact on Worst Group Accuracy (WGA):

\begin{itemize}
    \item \textbf{Perfect (Baseline):} We utilize the standard, fine-grained segmentation masks provided by the CUB dataset. This represents the ideal scenario where the object is perfectly isolated from its original background.
    
    \item \textbf{Noisy (Dilated):} To simulate automated segmentation masks that fail to tightly contour the object, we apply a morphological dilation (radius = 15px) to the original mask. This results in a ``halo'' effect, where the alignment model is exposed to the bird along with a significant margin of pixels from the original, potentially spurious, background.
    
    \item \textbf{Botched (Eroded):} To simulate over-aggressive filtering or partial occlusion, we apply a morphological erosion (radius = 21px) to the mask. This removes peripheral features of the bird (e.g., beaks, tails, and crests), forcing the model to align based on incomplete semantic information.
    
    \item \textbf{Bounding Box:} We replace the fine-grained mask with a rectangular bounding box derived from the mask's extents. This represents the coarsest possible segmentation, where the object is pasted along with all local background context contained within the box, serving as a lower-bound baseline for segmentation precision.
\end{itemize}

By comparing performance across these variations, we aim to determine if our method requires pixel-perfect isolation of the target object or if it remains effective even when the synthesis process includes spurious background noise or incomplete object features.

\begin{table}[h]
    \centering
    \caption{Sensitivity Analysis of Segmentation Quality. We report Average Accuracy and Worst Group Accuracy (WGA) across different segmentation degradation modes (mean $\pm$ std over 3 runs).}
    \label{tab:seg_ablation}
    \begin{tabular}{lcc}
        \toprule
        \textbf{Segmentation Mode} & \textbf{Average Accuracy} & \textbf{Worst Group Accuracy} \\
        \midrule
        \textbf{Perfect}  & $96.3 \std{0.8}$ & $92.5 \std{1.1}$ \\
        \textbf{Noisy}    & $95.3 \std{0.5}$ & $90.4 \std{1.7}$ \\
        \textbf{Botched}  & $91.3 \std{0.9}$ & $90.92 \std{1.4}$ \\
        \textbf{Bounding Box}     & $87.5 \std{1.6}$ & $78.9 \std{3.3}$ \\
        \bottomrule
    \end{tabular}
\end{table}

The results of this ablation study conducted for SigLIP 2, detailed in Table \ref{tab:seg_ablation}, reveal a clear hierarchy of performance correlated with segmentation fidelity, yet demonstrate a surprising degree of robustness. As expected, the \textbf{Perfect} segmentation yields the highest Worst Group Accuracy (91.80\%), serving as our upper bound. Notably, the \textbf{Botched} (eroded) variation outperforms the \textbf{Noisy} (dilated) variation in WGA (90.52\% vs. 89.98\%), with significantly lower variance. This suggests that for robust alignment, it is preferable to sacrifice peripheral object details (via erosion) rather than risk including spurious background features (via dilation). Finally, while the \textbf{Bounding Box} approach suffers a significant drop in performance (82.53\% WGA), it remains functional, indicating that the method can still extract useful semantic alignment signals even from coarse, unsegmented localization data, though removal of the original background remains critical for optimal performance.

\section{BAPs Impact on Scene Recognition and Limitations Thereof}\label{app: bap zero shot}

To understand how BAP affects the model's broader capabilities, we designed a zero-shot evaluation framework. This allowed us to compare the original, pre-trained CLIP vision encoder directly against our aligned version without the need for additional fine-tuning or linear probes. Our goal was to assess performance across two distinct areas: domain-specific object classification (Waterbirds test set) and open-domain scene classification (Places365 backgrounds).

\subsection{Zero-Shot Classifier Construction}

For all zero-shot tasks, we froze the text encoder and generated classifier weights using a standard prompt ensembling technique. Rather than relying on a single text prompt, we created a robust "prototype" for each class by inserting the class name into 40 distinct templates (e.g., ``a photo of a \{ \}'', ``a bad photo of a \{ \}''). 

We encoded these 40 variations and averaged their embeddings to create a single representative vector for each class. During inference, classification was performed by comparing the image embedding to these class prototypes; the model simply predicted the class whose prototype had the highest similarity score with the image. We performed this procedure twice: once using the original CLIP vision encoder weights and once using our aligned weights.

\subsection{Evaluation Tasks}

We evaluated the models on two distinct datasets to measure both the improvement in object focus and the retention of background knowledge.

\subsubsection{Task A: Robust Object Classification (Waterbirds)}
The primary goal of this task was to assess whether the alignment allows the model to better separate the in-class images in a zero-shot capacity. Given our results in Section \ref{sec: waterbirds results}, we expect performance to be higher on the BAP model than the native CLIP model.

After applying BAP using the same CUB-Places365 composites used in Section \ref{sec: waterbirds results}, we evaluated the model on the official Waterbirds test set using a simple binary classification scheme (``landbird'' vs. ``waterbird'').

\subsubsection{Task B: Background Retention (Places)}

This task investigates BAPs effect on CLIPs  general-purpose, zero-shot classification abilities.  We have already demonstrated that, in the presence of the core foreground signal, BAP vision encoders ignore most background variation.  However, what we seek to understand with this test is BAP's effect on the vision encoders representations of backgrounds and out-of-class objects \textit{in the absence of core foreground items}.

We aggregated images from the Places365 dataset corresponding to the background categories used during training (e.g., \textit{ocean, attic, igloo}). We then performed multi-class classification to verify if the model could still correctly identify these scenes when explicitly queried. The Places365  Dataset offers a  comprehensive supply of both general background scenes (e.g., ocean, rainforest) and more object centered scenes (e.g., igloo, windmill).

\begin{table}[h]
\centering
\renewcommand{\arraystretch}{1.25} 
\caption{\textbf{Zero-Shot Classification Results} We compare the vision encoder's ability to classify in a zero-shot regime for both native CLIP and post BAP CLIP. We investigate performance both within the target class and on out-of-class scenes and objects in order to assess the impact of BAP on the vision encoder's ability to recognize and represent background information.}
\vspace{5 pt}
\label{tab:zeroshot_results}
\begin{tabular}{lcc}
\hline
\textbf{Class Name} & \textbf{Original CLIP Accuracy} & \textbf{BAP CLIP Accuracy} \\ \hline
\multicolumn{3}{l}{\textit{Task A: Waterbirds (Target Class)}} \\ \hline
Landbird & 0.76 & \textbf{0.94} \\
Waterbird & 0.68 & \textbf{0.75} \\ \hline
\multicolumn{3}{l}{\textit{Task B: Places365 (Out-of-class scenes and objects)}} \\ \hline
Abbey & \textbf{0.79} & 0.13 \\
Alley & \textbf{0.95} & 0.65 \\
Attic & \textbf{0.92} & 0.02 \\
Auditorium & \textbf{0.89} & 0.04 \\
Bar & \textbf{0.92} & 0.16 \\
Bridge & \textbf{0.92} & 0.57 \\
Canyon & \textbf{0.92} & 0.01 \\
Castle & \textbf{0.76} & 0.38 \\
Cemetery & \textbf{0.96} & 0.36 \\
Chalet & \textbf{0.92} & 0.45 \\
Classroom & \textbf{0.93} & 0.00 \\
Closet & \textbf{0.98} & 0.05 \\
Crevasse & \textbf{0.89} & 0.02 \\
Driveway & \textbf{0.83} & 0.11 \\
Engine Room & \textbf{0.96} & 0.07 \\
Iceberg & \textbf{0.85} & 0.24 \\
Igloo & \textbf{0.93} & 0.81 \\
Martial Arts Gym & \textbf{1.00} & 0.61 \\
Ocean & \textbf{0.81} & 0.01 \\
Pond & \textbf{0.90} & 0.41 \\
Rainforest & \textbf{0.79} & 0.05 \\
Shopfront & \textbf{0.96} & 0.10 \\
Sky & \textbf{0.69} & 0.60 \\
Windmill & \textbf{0.97} & 0.70 \\ \hline
\end{tabular}
\end{table}

The results in Table \ref{tab:zeroshot_results} are dramatic: we observe higher zero-shot classification scores for our target class (birds) using BAP, however,  we note a catastrophic collapse in the BAP encoders ability to classify out-of-class scenes and objects.  This is expected given the embedding space visualizations seen in Figure \ref{fig:umap}. 

Given that BAPs `raison d'être' is to induce background invariance and isolate model attention to the desired class objects, this behavior is again, expected and further points to our method successfully modifying the encoders inductive bias. By treating the background as a nuisance variable during the anchor generation phase, the model transitions from a context-dependent estimator to an intrinsic object recognizer

This renders the adoption of BAP in real-world scenarios a nuanced task-specific decision;  in instances where background invariance is of high practical utility, BAP makes an excellent selection. However, in instances where environmental context acts as a vital signal, BAP may compromise performance. Therefore, BAP is most effectively deployed in specialized scenarios where background invariance is a high-priority requirement for robustness. For instance, in wildlife monitoring and camera trap applications, BAP prevents the model from utilizing specific rocky terrains or forest clearings as "shortcuts" for species identification, ensuring that animals are recognized by their physical characteristics even when they move to novel environments. Similarly, in medical diagnostic imaging, BAP can induce invariance to scanner artifacts or healthy tissue textures, forcing the encoder to focus exclusively on critical clinical markers.

On the other hand, in tasks such as autonomous driving, the background provides the necessary spatial context to disambiguate objects; a red octagon only becomes a "stop sign" when located within a traffic scene. Likewise, in action recognition, the surrounding environment is often essential for full comprehension, as the same physical movement may be interpreted as "playing golf" or "hailing a taxi" based entirely on whether the background is a golf course or a city street. Ultimately, while the "catastrophic collapse" of background classification scores in our results confirms the successful induction of background invariance, it highlights that BAP is best suited for  tasks where isolating the target from its surroundings is the primary objective.

\section{Training Progression and Fine-Tuning Constraints}\label{app: fine tuning restrictions}

This section gives an account of the  progression of the alignment objective $\mathcal{L}_{align}$ during BAP as well as the behavior of the model on the downstream Waterbirds benchmark. 

As far as  the alignment objective is concerned,  initially, we see a sharp decline in loss during the first few epochs, which gradually levels out to approach an asymptote of around $0.05$. There is not much insight there beyond the fact that BAP is relatively stable  during training.  Due to this stable behavior,  early stopping on $\mathcal{L}_{align}$ is recommended to reduce  unneeded compute resources.

\begin{figure*}[ht]
    \centering
    \begin{minipage}{\textwidth}
        \centering
        \includegraphics[trim={0.5cm} {0.2cm} {0.2cm} {0.5cm}, clip, width=\linewidth]{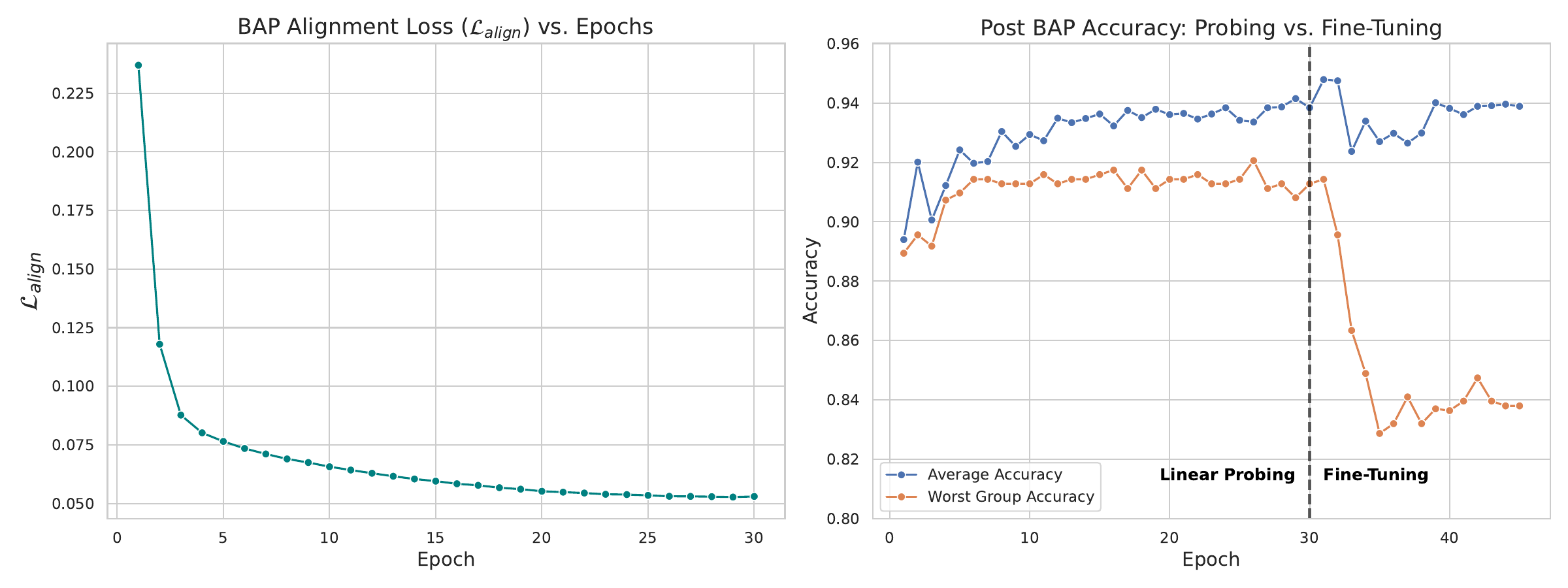}
            \caption{Left: progression of the alignment loss during BAP. Right:  progression of average and worst group accuracies during linear probing and then full fine-tuning post-BAP.  }
            \label{fig:loss progression}
\end{minipage}
\end{figure*}

As far as post-BAP dowstream accuracy is concerned,  we observe that peak performance is reached after only around 5 epochs of linear probe training. This is not unexpected as BAP produces well-organized, semantically grounded representations.  

Following 30 epochs of linear probe training, we unfreeze the weights of the vision encoder and fine-tune Using standard cross-entropy on the Waterbirds training set. We note that performance, particularly as regards worst group accuracy, degrades sharply after the weights of the model are fine-tuned on the  spuriously correlated Waterbirds training set. This places a critical restriction on downstream deployment since fine-tuning the vision encoder in this manner undoes BAP's induced background invariance. Therefore, we strongly recommend that any post-BAP downstream deployment  utilize the vision encoder as a fixed  feature extractor. Given the strong sim-to-real and OOD performance exhibited in Section \ref{sec: counteranimal},  and given that VLMs possess rich semantic representations, we do not anticipate this to be a major limitation of our method.

\section{Appendix: COCO gender classification  dataset construction }
\label{sec:appendix_coco_gender}

To construct the synthetic COCO gender classification dataset utilized in Section \ref{sec: results} (specifically Table \ref{tab:gender}), we required instance-level gender labels. Because the standard MS-COCO dataset provides only a generic \texttt{person} category, we developed an automated pipeline utilizing zero-shot Vision-Language Models to assign ``man'' and ``woman'' pseudo-labels. Crucially, to ensure the downstream classification task remained suitably challenging and avoided trivial, hyper-stereotypical edge cases, we implemented a bounded margin selection and flattened sampling strategy.

\subsection{Instance Isolation and Preprocessing}
We began by filtering the MS-COCO 2017 training set for \texttt{person} instances. To ensure sufficient visual detail for accurate zero-shot classification, we discarded any instance with a bounding box resolution smaller than $112 \times 112$ pixels. 

For each valid instance, the foreground subject was extracted using its corresponding ground-truth bounding box and instance segmentation mask. The mask was thresholded and the isolated foreground was resized to $224 \times 224$ pixels via Lanczos resampling, preserving the original aspect ratio. To remove any spurious background context during the labeling phase, the isolated subject was pasted onto a neutral gray canvas (RGB: 128, 128, 128). 

\subsection{Zero-Shot Classification and Prompt Ensembling}
To perform the zero-shot classification, we utilized a pre-trained OpenCLIP ViT-H-14 model (\texttt{laion2b\_s32b\_b79k}). To construct robust text prototypes for our target classes, we employed prompt ensembling. Specifically, rather than relying on a single text prompt, we embedded multiple descriptive templates (e.g., ``a photo of a man'', ``a portrait of a woman'') and averaged their normalized embeddings to create a single, robust prototype vector for both ``man'' and ``woman''.

Each preprocessed, background-free image tensor was passed through the vision encoder to obtain its $\ell_{2}$-normalized image embedding. We then computed the cosine similarity between the image embedding and the two text prototypes. These similarity scores were passed through a Softmax function to yield confidence probabilities. Finally, we calculated a classification margin $M$ for each instance, defined as:
\begin{equation}
    M = C_{man} - C_{woman}
\end{equation}
where $C_{man}$ and $C_{woman}$ denote the Softmax confidence probabilities for the respective classes. Consequently, $M \in [-1.0, 1.0]$, with positive values indicating a ``man'' prediction and negative values indicating a ``woman'' prediction.

\subsection{Bounded Margin Selection and Flattened Sampling}
Relying strictly on the absolute highest margin instances (e.g., $|M| \approx 1.0$) risks populating the dataset with overly simplistic, trivial, or hyper-stereotypical examples, which degrades the quality of the downstream robustness evaluation. To curate a dataset with a diverse spectrum of classification difficulties, we applied the following selection and sampling protocol:

\begin{itemize}
    \item \textbf{Bounded Selection:} We filtered the dataset to include only instances where the absolute margin fell within a specific confidence window. We defined our target bounds as $0.20 \leq |M| \leq 0.99$. This effectively excluded instances where the model was too uncertain (margin $< 0.20$) and instances that were overly simplistic (margin $> 0.99$).
    \item \textbf{Flattened Distribution Sampling:} Because the natural distribution of margins is heavily skewed towards high confidence values, taking a random sample from the bounded pool would result in a dataset dominated by high confidence instances. To enforce a uniform difficulty distribution, we stratified the bounded pool into 15 equally spaced bins between the minimum (0.20) and maximum (0.99) absolute margins.
    \item \textbf{Round-Robin Extraction:} We performed a randomized round-robin sampling across these 15 bins, iteratively drawing one instance from each bin until we reached our target dataset size.
\end{itemize}

\subsection{Final Dataset Splits}
Using this flattened sampling methodology, we extracted 4,000 instances for the ``man'' class and 4,000 instances for the ``woman'' class. These sets were subsequently randomized and divided into distinct training and testing splits, yielding 3,000 training instances and 1,000 testing instances per class. This carefully curated subset ensures a balanced, sufficiently challenging evaluation benchmark for the experiments detailed in the main body.

\subsection{Spurious Dataset Construction and Compositing}
\label{sec:appendix_coco_compositing}

Following the extraction of  ``man'' and ``woman'' foreground instances, we explicitly constructed a dataset exhibiting severe, domain-specific spurious correlations. We designed this dataset to mimic real-world demographic biases by placing instances into stereotypical contexts using backgrounds sourced from the Places365 dataset.

\subsubsection{Background Selection and Spurious Correlation Injection}
We established two distinct pools of target backgrounds to serve as our spurious features:
\begin{itemize}
    \item \textbf{Male-Correlated Backgrounds:} Images drawn from the \texttt{excavation} and \texttt{construction\_site} categories.
    \item \textbf{Female-Correlated Backgrounds:} Images drawn from the \texttt{kitchen} and \texttt{kitchenette} categories.
\end{itemize}

We configured the training distribution with either a 95\% or 100\% spurious correlation rate.  Conversely, for the generation of background-invariant anchor vectors (Phase 1 of BAP), we utilized a separate, highly diverse pool of 50 unconfounded background categories (e.g., \texttt{coast}, \texttt{alley}, \texttt{classroom}, \texttt{rainforest}). This ensured the model learned to decouple the foreground from a wide variety of contexts before encountering the strictly correlated downstream task.

\subsubsection{Image Compositing Pipeline}
To seamlessly integrate the isolated COCO foregrounds into the Places365 backgrounds, we implemented a deterministic compositing pipeline designed to minimize edge artifacts that the network might exploit as trivial classification shortcuts. The process for each target resolution of $224 \times 224$ was defined as follows:

\begin{enumerate}
    \item \textbf{Scaling:} The isolated foreground object and its corresponding mask were proportionally downscaled to occupy a maximum of 75\% of the target background dimensions ($168 \times 168$ pixels) using Lanczos resampling.
    \item \textbf{Mask Smoothing:} To prevent sharp, jagged boundaries between the foreground and the new background, the segmentation mask was strictly thresholded (pixel values $> 100$ mapped to 255) and subsequently filtered using a Gaussian blur with a radius of $\sigma=1$.
    \item \textbf{Centering:} The preprocessed foreground and mask were pasted directly into the center offset of the chosen $224 \times 224$ background image.
\end{enumerate}

\subsubsection{Evaluation Split}
While the primary training set enforced a high or perfect spurious correlation, the evaluation required a strictly balanced test set to calculate worst-group accuracy accurately. To achieve this, the test set was constructed to uniformly represent all four possible gender-background combinations: Man/Construction, Man/Kitchen, Woman/Construction, and Woman/Kitchen. Furthermore, to prevent the model from memorizing specific background instances, the background images themselves were divided into an 80/20 train/test split, ensuring that the scenes encountered during evaluation were strictly unseen during both alignment and linear probing phases.

The code and  and metadata required to recreate the training and testing versions of this data set are included in the supplemental material.

\subsection{Disclaimer on Automated Gender Pseudo-Labeling}

We explicitly acknowledge the limitations and ethical considerations inherent in utilizing a VLM (OpenCLIP ViT-H-14) to assign instance-level gender pseudo-labels. These labels strictly represent visual predictions based on the model's learned associations of gender presentation, rather than true, self-identified demographic data. 

To mitigate the risk of circularity—whereby the VLM evaluating the bias is tested on labels inherently confounded by that same bias—we implemented two structural safeguards. First, to prevent background-induced label leakage, all pseudo-labels were generated using exclusively isolated, segmented foreground subjects placed against neutral grey canvases. By completely stripping the original background context prior to zero-shot inference, we successfully decoupled the labeling mechanism from the specific spurious environmental correlations (e.g., kitchens, construction sites) being evaluated in the downstream task. Second, we utilized a larger, higher-capacity teacher model (ViT-H-14) to generate the baseline labels for the evaluations conducted on the base-level student architectures (ViT-B-16), leveraging the teacher's stronger representational capacity to ensure higher fidelity visual pseudo-labels.

Consequently, this dataset is strictly formulated and deployed as a controlled stress-test of a vision encoder's reliance on spurious spatial context and background correlations, rather than a definitive measure of real-world human demographic bias.

\newpage
\section{Appendix:  implementation and experimental Details }\label{app:implementation deets}

\subsection{Model Initialization and Backbone Selection}
\label{sec:model_init}

To ensure complete reproducibility, all experiments utilize specific, publicly available pre-trained checkpoints rather than default library initializations. For our foundational Softmax-based baseline, we employ the CLIP ViT-B/16 architecture. We initialize this model using the \texttt{laion2b\_s34b\_b88k} weights (trained on the LAION-2B dataset) provided by the OpenCLIP library \citep{laion}. This model was used in the evaluations presented in Tables \ref{tab:waterbirds}, \ref{tab:gender}, \ref{tab:counteranimal results}; to demonstrate cross-architectural stability and transferability from transformer-based backbones to convolutional backbones, we utilize a CLIP ConvNeXT (trained on laion2b\_s13b\_b82k) backbone for the results presented in Tables \ref{tab:nico++ results} and \ref{tab:waterbirds_convnext}.  

For our state-of-the-art Sigmoid-based evaluations, we utilize SigLIP 2. Specifically, we load the \texttt{google/siglip2-base-patch16-224} Hugging Face weights. To ensure parity across evaluations, all input images for both architectures are strictly resized to 224x224 pixels and normalized using standard CLIP mean and standard deviation statistics. For backbone size ablations and non-LAION CLIP checkpoint evaluations, see Appendix \ref{app:backbone_ablation}.

\subsection{Implementation Details: External Baselines}
\label{sec:external_baselines_impl}

To ensure a rigorous and fair comparison, all external baselines were implemented faithful to their original formulations, utilizing the same underlying feature extractors and data loaders as our proposed method. Where applicable, baseline-specific hyperparameters were tuned using the validation splits to maximize Worst-Group Accuracy (WGA).

\textbf{RoboShot:} We implement the zero-shot robustification algorithm using predefined harmful and helpful concept pairs. For instance, in the Waterbirds domain, harmful vector projections (e.g., ``water background'' vs. ``land background'') are analytically subtracted from the image embeddings, while helpful projections (e.g., ``waterbird'' vs. ``landbird'') are added. Predictions are subsequently generated via cosine similarity with the zero-shot text head \citep{roboshot}.

\textbf{DIAL (Disentangle, Identify, And Label-free removal):} We utilize a Matryoshka Sparse Autoencoder (MSAE) \citep{sae} with a latent dimension expanded by a factor of 8 relative to the backbone embedding ($d_{latent} = 8 \times d_{model}$). Following the optimal parameters established in prior work, we set the selection fraction $\alpha = 0.65$, projection scaling $\lambda = 0.9$, and $k = 10$ for the K-Nearest Neighbors candidate selection. Spurious feature attribution is guided by target text prompts (e.g., ``water background'').\citep{dial}

\textbf{WiSE-FT (Weight-Space Ensembles for Fine-Tuning):} We first train an ERM feature extractor using a Linear Probe followed by Fine-Tuning (LP-FT) paradigm. The resulting weights are then linearly interpolated with the original zero-shot weights of the model. We utilize an interpolation coefficient of $\alpha = 0.5$ across all experiments.\citep{wiseft}

\textbf{PruSC:} This representation-level intervention operates in four phases. First, a dense ERM baseline is trained. Second, embeddings are clustered using K-Means ($k=8$), and a balanced task dataset ($D_{task}$) is sampled at a ratio of 0.1. Third, MaskedLinear layers are introduced and trained for 15 epochs (Adam optimizer, learning rate $1\times 10^{-2}$) utilizing a contrastive loss ($\beta = 1.0$) and sparsity penalty ($\alpha_{sparse} = 1\times 10^{-4}$). Finally, the learned masks are binarized and frozen, while the active subnetwork weights are fine-tuned for 30 epochs.\citep{prusc}

\textbf{DFR (Deep Feature Reweighting):} We extract features from the validation set using a fully trained ERM model. A Logistic Regression head with an L1 penalty is tuned over $C \in \{1.0, 0.7, 0.3, 0.1, 0.07, 0.03, 0.01\}$ using a 50\% split of the validation data to maximize WGA. To ensure stability, the final classification head is retrained 10 times on randomly sampled, group-balanced subsets of the remaining validation data, and the resulting weights and biases are averaged.\citep{dfr}

\textbf{AFR (Automatic Feature Reweighting):} The training data is split into an 80\% ERM training set and a 20\% reweighting set. After training the base ERM model for 5 epochs, the $\gamma$ penalty parameter is tuned via grid search ($\gamma \in \{0, 1, 3, 6, 10, 20\}$) using the validation set WGA. The final classification head is retrained for 100 iterations using the optimal $\gamma$ to dynamically down-weight majority group samples.\citep{afr}

\subsection{Waterbirds Benchmark: Internal Baselines and Experimental Setup}
\label{sec:waterbirds_setup}

All experiments on the Waterbirds benchmark utilize a unified data processing pipeline. Images are resized to $224 \times 224$ and normalized using standard CLIP/SigLIP 2 statistics. Training subsets employ random horizontal flipping, while evaluation subsets utilize standard center cropping. 

\subsubsection{Internal Baselines}\label{sec:internal_baselines}

\textbf{Zero-Shot (Native ZS):} Zero-shot classification is performed by ensembling text prompts across 40 distinct templates. The encoded text vectors are $L_2$-normalized, averaged to form a single prototype vector per class, and re-normalized. 

\textbf{ERM Linear Probe (ERM LP):} A linear classification head is trained on top of the frozen vision backbone for 30 epochs using the AdamW optimizer (learning rate $5\times 10^{-4}$). To account for dataset imbalance, we apply a weighted Cross-Entropy loss inversely proportional to class frequencies.

\textbf{ERM LP-FT:} We employ a two-stage stabilization strategy as outlined in \citep{lp_then_ft}. The linear head is first probed for 5 epochs (AdamW, learning rate $1\times 10^{-3}$). Subsequently, the backbone is unfrozen, and the entire model is fine-tuned for 30 epochs. We use a Sequential Learning Rate scheduler consisting of a 200-step linear warmup (from 0.01 to 1.0) followed by Cosine Annealing decay down to $1\times 10^{-6}$.

\textbf{Data-Matched ERM Control:} To causally isolate the effect of our proposed alignment loss, this control utilizes the exact same synthetic composite generation pipeline and virtual epoch scaling as our method. The model is trained via standard Cross-Entropy for 30 epochs, followed by a final linear probe (identical to the ERM LP setup above) stage on real data to emulate the exact data exposure and computational budget of BAP.

\textbf{Background-Invariant Anchor Pre-training (BAP, Ours):} Our method begins by sampling $N=5,000$ class-balanced instances from the CUB dataset. In Phase 1, each instance is composited onto 10 random Places365 backgrounds to distill a mean-pooled, $L_2$-normalized target anchor. In Phase 2, the backbone aligns the student embeddings to these anchors using a Cosine Embedding Loss for 30 epochs (AdamW, learning rate $5\times 10^{-6}$, weight decay $0.01$). Finally, Phase 3 executes either a standard linear probe on the target domain or Zero-Shot classification using the same configurations as the (ERM LP and Zero-Shot) specified above.

\subsubsection{Dataset Construction and Leakage Prevention}
\label{sec:dataset_construction}

\subsubsection{Data Leakage Prevention}
When constructing synthetic datasets for robustness pre-training, ensuring zero overlap with downstream validation or test sets is paramount. For the Waterbirds benchmark, we construct our synthetic training set using a subset of approximately 5,000 samples from the CUB dataset. 

To rigorously prevent data leakage, we cross-reference the file paths of our CUB instances against the official Waterbirds metadata. Any CUB image that appears in either the Waterbirds validation or test split is explicitly filtered out and excluded from the BAP anchor generation and alignment phases. Consequently, the visual encoder is never exposed to downstream evaluation images during pre-training.

\paragraph{Synthetic Compositing Details}
During BAP Phase 1 (Anchor Extraction) and Phase 2 (Alignment), foreground objects are superimposed onto 50 distinct background categories sourced from Places365. We employ a deterministic compositing function to ensure consistency:

\begin{enumerate}
    \item \textbf{Isolation:} The foreground bird is isolated using its provided segmentation mask and cropped tightly to its bounding box.
    \item \textbf{Scaling:} The isolated foreground and its mask are scaled to a random number in the range $[0.6-0.8]$ as a fraction of the target 224x224 resolution using Lanczos resampling.
    \item \textbf{Mask Refinement:} To prevent background noise from bleeding through transparent or soft edges, the mask is solidified via a hard threshold (pixel values > 100 are set to 255). Subsequently, a Gaussian blur with a radius of 1 is applied to smooth the composite edges.
    \item \textbf{Placement:} The resized foreground is centered and pasted onto the target 224x224 background image using the refined mask.
\end{enumerate}

\subsubsection{BAP Hyperparameters and Optimization Strategy}

Table \ref{tab:bap_hyperparams} details the specific hyperparameter configuration utilized during Phase 2 (Alignment Pre-training) of BAP. 

Rather than relying on an exhaustive and computationally expensive grid search, hyperparameters were established through targeted empirical tuning, largely inheriting standard defaults utilized for fine-tuning Vision Transformers. We observed that the $\mathcal{L}_{align}$ objective is remarkably stable and not highly sensitive to precise hyperparameter selections. The ability to achieve near-peak Worst-Group Accuracy (WGA) without rigorous, dataset-specific hyperparameter sweeps underscores the robustness of the BAP algorithm and its suitability for practical, out-of-the-box deployment.

\begin{table}[h]
\centering
\small
\caption{Hyperparameter configuration for BAP Phase 2 (Alignment Pre-training).}
\label{tab:bap_hyperparams}
\begin{tabular}{ll}
\toprule
\textbf{Hyperparameter} & \textbf{Value} \\
\midrule
Loss Function & Cosine Embedding Loss (Margin $1.0$) \\
Optimizer & AdamW \\
Learning Rate (Backbone) & $5 \times 10^{-6}$ \\
Weight Decay & 0.01 \\
Batch Size & 128 \\
Training Epochs & 30 \\
Learning Rate Scheduler & Linear Warmup (10\%) $\rightarrow$ Cosine Annealing \\
$N$ (number of  unique foreground instances ) & 5000 \\
$M$ ( number of contexts each foreground is placed on ) & 5 \\
$K$ ( used in anchor vector construction ) & 10 \\
Mixed Precision & bfloat16 \\
\bottomrule
\end{tabular}
\end{table}

The learning rate scheduler utilizes a linear warmup over the first 10\% of the total training steps to prevent early gradient spikes, followed by a standard cosine decay.

\subsubsection{Statistical Significance and Variance}\label{app:random seeds explained}
To ensure robust performance estimates, all routines are evaluated across $N=5$ independent, sequential runs. Because the global random seed governs the sequence generation, the random state progresses natively across runs. Consequently, variance across runs captures the compounding effects of (1) differing linear head initializations, (2) variations in dataset shuffling, and (3) the stochastic pairings of foreground objects and background scenes during synthetic composite generation. We report the mean and standard deviation for Average Accuracy and Worst-Group Accuracy (WGA). Note that this applies to all other results presented on other datasets.

\subsubsection{Compute Resources and Efficiency}
All experiments were executed on a single NVIDIA L40S GPU using PyTorch with `bfloat16` mixed precision. To overcome critical Disk I/O bottlenecks during synthetic composite generation, we implemented an LRU-cached image loading system, keeping up to 20,000 frequently accessed backgrounds in memory. A single full pipeline run (evaluating all 11 routines across both CLIP and SigLIP2 variants) requires approximately 12 GPU hours.

\subsection{COCO Gender Benchmark: Experimental Setup and Baseline Specifics}
\label{sec:coco_gender_setup}

The COCO Gender classification benchmark evaluates model robustness against severe demographic spurious correlations (i.e., gender correlated with stereotypical background scenes). The dataset construction, object isolation, and compositing pipeline are detailed in Appendix \ref{sec:appendix_coco_gender}. Unless otherwise stated, the foundational model initialization (CLIP, SigLIP 2), hardware constraints, and standard optimization hyperparameters (e.g., learning rates, AdamW weight decay, batch sizes, and 30-epoch training schedules) are identical to those outlined in the Waterbirds setup (Section \ref{sec:waterbirds_setup}). 

\subsubsection{Task Formulation and Metrics}
Models are evaluated on a 4-way group robustness task: Male on Male-correlated backgrounds (Construction/Excavation), Male on Female-correlated backgrounds (Kitchen), Female on Male-correlated backgrounds, and Female on Female-correlated backgrounds. We report the Average Accuracy and Worst-Group Accuracy (WGA) across $N=5$ independent runs with randomized dataset shuffling and compositing permutations.

\subsubsection{Baseline Implementations and Task-Specific Adjustments}

\textbf{Zero-Shot and Text Prompts:} For the Zero-Shot baseline and the text-target generation in our proposed method, we utilize a modified set of prompt templates better suited for human subjects: \textit{``a photo of a \{\}.''}, \textit{``an image of a \{\}.''}, and \textit{``a picture of a \{\}.''}. The class tokens are explicitly mapped to \textit{``male person''} and \textit{``female person''}.

\textbf{BAP (Ours) Alignment Parameters:} The Background-Invariant Anchor Pre-training (BAP) proceeds similarly to the Waterbirds experiment. During Phase 1 (Distillation), target vectors are derived by averaging the embeddings of each instance pasted across 10 randomly sampled, unconfounded \texttt{Places365} scenes. Hyperparameters and setup are identical to those  specified in Table \ref{tab:bap_hyperparams}.

\textbf{DFR Validation Splitting:} Deep Feature Reweighting (DFR) requires a group-balanced validation set to tune the $L_1$ penalty parameter ($C$) and retrain the final classification head. Because our primary COCO training split enforces a 100\% spurious correlation (meaning minority groups are entirely absent), a standard validation split cannot be used. Instead, we explicitly partition 100 samples per group from the initial balanced held-out test set to serve strictly as the DFR calibration and tuning split. All reported final metrics for DFR (and all other methods) are computed exclusively on the remaining, strictly unseen evaluation samples.

\subsection{CounterAnimal Benchmark: Experimental Setup and Downstream Adaptation}
\label{sec:CounterAnimal_setup}

The CounterAnimal benchmark (Table \ref{tab:counteranimal results}) evaluates the sim-to-real transferability of our method. It assesses whether the robustness gained from synthetic composites (CUB objects on Places365 backgrounds) transfers to natural, real-world images exhibiting severe distribution shifts and 100\% spurious correlations.

\subsubsection{Pre-training (BAP and Data-Matched Control)}
The pre-training phase for both BAP and the Data-Matched ERM Control remains identical to the setup described in the Waterbirds experiments (Section \ref{sec:waterbirds_setup} and Table \ref{tab:bap_hyperparams}). Both methods utilize the same synthetic dataset of CUB foregrounds composited onto Places365 backgrounds, sharing the exact same epoch size, learning rates ($5 \times 10^{-6}$), and optimizers. This ensures that any performance differences on the CounterAnimal downstream tasks are attributable strictly to the objective function (Cosine Embedding Alignment vs. standard Cross-Entropy) rather than the pre-training data distribution.

\subsubsection{Downstream Data Preprocessing: Deterministic Smart Cropping}
Unlike the tightly standardized CUB or COCO instances, CounterAnimal consists of natural photographs with highly variable aspect ratios. To prevent standard random cropping from inadvertently removing the target animal or destroying the background context, we implemented a deterministic ``Smart Crop'' strategy:
\begin{itemize}
    \item \textbf{Landscape Images ($W > H$):} 50 pixels are cropped symmetrically from the left and right edges.
    \item \textbf{Portrait Images ($H > W$):} 50 pixels are cropped symmetrically from the top and bottom edges.
\end{itemize}
Following the crop, the images are resized to the target $224 \times 224$ resolution using Bicubic interpolation with anti-aliasing enabled, followed by standard normalization. Random horizontal flipping is applied exclusively during training.

\subsubsection{Downstream Probing, Fine-Tuning, and Evaluation}
For the downstream binary classification tasks (e.g., Ptarmigan vs. Prairie-Chicken, Brambling vs. Bulbul), the dataset sizes per pair are relatively small. To account for this, we make the following adjustments to the downstream adaptation protocol:

\textbf{Optimization and Batching:} The batch size is reduced to 32. Linear Probing (LP) is conducted for 40 epochs using the AdamW optimizer with a learning rate of $1 \times 10^{-4}$. For Fine-Tuning (FT) comparisons, the backbone learning rate is strictly constrained to $1 \times 10^{-6}$ while the head is trained at $1 \times 10^{-4}$ for 40 epochs. To account for dataset imbalance, we apply a weighted Cross-Entropy loss inversely proportional to class frequencies.

\subsubsection{CounterAnimal Dataset Construction and Task Formulation}
\label{sec:counteranimal_construction}

Unlike our synthetic pre-training distributions, CounterAnimal consists of unedited, natural photographs where animals are captured in both common (stereotypical) and counter (atypical) background contexts. 

For our downstream adaptation, we formulate binary classification tasks using visually similar animal pairs that share overlapping taxonomic orders or camouflage characteristics (e.g., Ptarmigan vs. Prairie-Chicken). To rigorously stress-test the models against severe distribution shifts, we explicitly construct the downstream training and testing splits to exhibit a perfect (100\%) spurious correlation:

\begin{itemize}
    \item \textbf{Correlated Training Split:} The training dataset is assembled such that each animal class appears exclusively within a single background context. For instance, in a given binary task, all instances of Class A are drawn from Context A (e.g., snow), and all instances of Class B are drawn from Context B (e.g., grass). A random 20\% subset of this strictly correlated data is reserved as a test split.
    
    \item \textbf{Counter-Context Evaluation Split:} To measure true background-invariant robustness and out-of-distribution (OOD) generalization, the held-out test set consists entirely of the inverted pairings along with the reserved correlated 20\% split.
\end{itemize}

This evaluation setup guarantees that any model relying on background features as a classification shortcut during the linear probing or fine-tuning phase will fail catastrophically on the test set. Success on this counter-context split strictly requires the visual backbone to have retained the robust, foreground-centric representations distilled during the BAP pre-training phase.

\textbf{Statistical Variance:} Due to the limited size of the downstream datasets, performance can be highly sensitive to the initial train/validation splits. To ensure rigorous estimates of statistical significance (Checklist Item 7), we execute $N=10$ independent runs using a fixed set of highly diverse global random seeds. These seeds explicitly govern the random 80/20 train-test splits, the weighted sampler stochasticity, and the classification head initializations.

\subsection{NICO++ Benchmark: Pre-training and Experimental Setup}
\label{sec:nico_setup}

The NICO++ vehicle classification tasks evaluate BAP's generalization capabilities using complex, real-world objects. Due to the substantial scale of the pre-training dataset (around $N= 80,000$ unique foreground instances compared to all previous datasets which had around $N = 5,000$ ), the backbone pre-training phase was decoupled from downstream testing.

\subsubsection{COCO Vehicle Dataset Construction}
The pre-training dataset was synthesized using isolated vehicle instances extracted from the MS-COCO 2017 training split. We explicitly filtered the dataset for six vehicle super-classes: \textit{car, bus, airplane, boat, train,} and \textit{bike}. Note we explicitly withheld \textit{truck} and \textit{motorbike} instances to test generalization at the superclass level.  We note a minor discrepancy in the class labels used throughout the paper: while COCO uses \textit{bicycle} and \textit{motorcycle}, we abbreviate these labels to \textit{bike} and \textit{motorbike}, respectively, solely to conserve space in the already dense Tables \ref{tab:nico++ results} and \ref{tab:nico_vit}. The underlying data and class definitions remain unchanged.

To ensure sufficient semantic detail for the vision encoder, any instance with a bounding box resolution smaller than $28 \times 28$ pixels was discarded. The isolated foreground objects were composited onto a massive, diverse pool of background images sampled from both the Places365 database and the Describable Textures Dataset (DTD), ensuring extreme background variance during alignment. 

\subsubsection{Pre-training Hyperparameters (BAP and ERM Control)}
To accommodate the larger dataset size, all pre-training was executed on a single NVIDIA H100 GPU using \texttt{bfloat16} mixed precision, with a batch size of 512. Both BAP and the Data-Matched ERM Control were trained for a fixed duration of 160 epochs to guarantee convergence.

\textbf{BAP Alignment:} During Phase 1, instance-specific anchor vectors were distilled by averaging the embeddings of each COCO vehicle across 10 randomly sampled backgrounds ($M =10$). In Phase 2, the backbone was optimized using the Cosine Embedding Loss via the AdamW optimizer (weight decay $\lambda = 0.01$, learning rate $7 \times 10^{-6}$). The learning rate schedule utilized a 400-step linear warmup followed by Cosine Annealing down to a minimum of $5 \times 10^{-7}$. 

\textbf{Data-Matched ERM Control:} The ERM baseline utilized the identical COCO-Places365/DTD dataset but was optimized using standard Cross-Entropy. To maintain stability, it followed a two-stage process: a 10-epoch linear probe on frozen representations (Head LR $1 \times 10^{-3}$), followed by 160 epochs of full fine-tuning. During fine-tuning, the backbone learning rate was constrained to $7 \times 10^{-6}$ while the head was optimized at $5 \times 10^{-4}$, utilizing the same 200-step warmup and cosine decay scheduler.

All other external and internal baselines follow the setup in \ref{sec:external_baselines_impl} and \ref{sec:internal_baselines} respectively.
  
\subsubsection{Downstream Adaptation and Evaluation}
Following the 160-epoch pre-training phase, the frozen backbones were evaluated on the real-world NICO++ vehicle classification pairs under a 100\% spurious correlation regime (e.g., all training cars on grass, all training trucks on water, with a 20\% split reserved for testing similar to the CounterAnimal setup \ref{sec:counteranimal_construction} (along of course with the swapped background sets).

The downstream adaptation protocol—including the 80/20 train-test splits, the utilization of a \texttt{WeightedRandomSampler} to handle class imbalance, the 40-epoch Linear Probing/Fine-Tuning duration, and the 5-run statistical variance tracking—is strictly identical to the methodology detailed for the CounterAnimal benchmark in Section \ref{sec:CounterAnimal_setup}.

Training a single instance of the model on an NVIDIA H100 using the full dataset requires approximately 6 hours. Given that our experiments consist of five independent runs for each configuration, evaluated across two baselines (BAP and data matched control) and two model variants (CLIP ConvNeXT and SigLIP 2), reproducing the complete set of experiments would require roughly $6 \times 5 \times 2 \times 2 = 120$ GPU hours in total.



\end{document}